\documentclass[preprint,10pt,authoryear]{elsarticle}

\setcitestyle{numbers,square,comma}
\usepackage{enumerate}
\usepackage{enumitem}
\usepackage{xspace}
\usepackage{gensymb}
\usepackage{textcomp}
\usepackage[nottoc]{tocbibind}
\usepackage{bm}
\usepackage{comment}
\usepackage{caption}
\usepackage{algorithm}% <=================http://ctan.org/pkg/algorithms
\usepackage{algpseudocode}

\usepackage{float}
\usepackage[hidelinks]{hyperref} 

\usepackage[a4paper, total={6in, 8in}]{geometry}
\usepackage{amssymb,amsmath}
\usepackage{diagbox}
\usepackage{longtable,booktabs,tabularx,tabulary}
\usepackage{multirow}
\usepackage{amsmath} % for "gathered" env.
\usepackage{rotating}
\usepackage{siunitx}
\usepackage[skip=0.33\baselineskip]{caption}
\usepackage{caption} 
\usepackage{amsmath}

\newcommand{\icol}[1]{% inline column vector
  \left[\begin{matrix}#1\end{matrix}\right]%
}

\journal{Elsevier}

\begin{document}

\begin{frontmatter}



\title{Physics-based Reduced Order Modeling for Uncertainty Quantification of Guided Wave Propagation using Bayesian Optimization}

\author[a]{G. I. Drakoulas}\ead{g.drakoulas@upnet.gr}
\author[a]{T. V. Gortsas\corref{cor}}\ead{gortsas@upatras.gr}
\author[a]{D. Polyzos}\ead{polyzos@mech.upatras.gr}

\cortext[cor]{Corresponding author}

\affiliation[a]{Department of Mechanical Engineering and Aeronautics, University of Patras,
Patras, GR-26500, Rion, Greece}

\begin{abstract}
In the context of digital twins, structural health monitoring (SHM) constitutes the backbone of condition-based maintenance, facilitating the interconnection between virtual and physical assets. Guided wave propagation (GWP) is commonly employed for the inspection of structures in SHM. However, GWP is sensitive to variations in the material properties of the structure, leading to false alarms. In this direction, uncertainty quantification (UQ) is regularly applied to improve the reliability of predictions. Computational mechanics is a useful tool for the simulation of GWP, and is often applied for UQ. Even so, the application of UQ methods requires numerous simulations, while large-scale, transient numerical GWP solutions increase the computational cost. Reduced order models (ROMs) are commonly employed to provide numerical results in a limited amount of time. In this paper, we propose a machine learning (ML)-based ROM, mentioned as \textit{BO-ML-ROM}, to decrease the computational time related to the simulation of the GWP. The ROM is integrated with a Bayesian optimization (BO) framework, to adaptively sample the parameters for the ROM training. The finite element method is used for the simulation of the high-fidelity models. The formulated ROM is used for forward UQ of the GWP in an aluminum plate with varying material properties. To determine the influence of each parameter perturbation, a global, variance-based sensitivity analysis is implemented based on Sobol' indices. It is shown that Bayesian optimization outperforms one-shot sampling methods, both in terms of accuracy and speed-up. The predicted results reveal the efficiency of \textit{BO-ML-ROM} for GWP and demonstrate its value for UQ.
\end{abstract}

\begin{keyword}
Guided waves \sep Reduced order modeling \sep Machine learning  \sep Uncertainty quantification \sep Bayesian optimization \sep Structural health monitoring

\begin{comment}
    The hybridization of numerical modeling with artificial intelligence techniques can significantly contribute to the design of physics-based digital twins for real-world applications, since the simulation time for physical systems can be significantly reduced. 
\end{comment}
\end{keyword}

\end{frontmatter}


\section{Introduction}
In the frame of the fourth industrial revolution, artificial intelligence, the internet of things, edge computing, and mixed reality are key technologies, enabling the implementation of digital twins (DTs) \cite{vettori2023adaptive, ritto2021digital}. Using real-world data combined with computational modeling and simulation, the DT links the physical with the digital world, resulting in a powerful tool for the virtual sensing of the physical asset \cite{thelen2022comprehensive, eleftheroglou2018structural}. In this direction, structural health monitoring (SHM) serves to assist in the inspection of the structures using condition-based maintenance, leading to improved life-cycle management \cite{Kopsaftopoulos, saeedifar2020damage, nastos2023non}. Through the installation of distributed sensors (e.g., accelerometers), SHM aims to track the relationship between changes made to the system characteristics (crack propagation, material degradation, etc.) and the operational performance of the structures \cite{janapati2016damage, Amer-Kopsaftopoulos19-VFS}. To achieve that, data-driven modeling techniques are implemented in the SHM systems to detect and identify damages \cite{su2006guided, rautela45simulation}. SHM methods have been employed in various types of structures, i.e., aircraft wings and panels in aerospace, bridges, and buildings in civil engineering, offshore wind farms in the energy sector, as well as offshore oil and gas platforms \cite{koune2023bayesian}. They are mainly classified as active or passive, depending on whether actuation or excitation signals are used to actively extract information about the condition of the structure. Inspection of the structure using guided wave propagation (GWP) is one of the most important techniques used in active sensing systems, since it does not require intricate or expensive equipment \cite{bellam2021parametric}. GWP can be used locally to inspect a small area or globally to investigate the behavior of the entire structure \cite{ahmed2021stochastic, nastos20182d}. 

In the context of nondestructive tests, GWP is commonly employed in the SHM community for damage detection, localization, and characterization, by analyzing the symmetric and antisymmetric modes of GWP \cite{ahmed2021stochastic, rekatsinas2015time}. The structural response strongly depends on the operational conditions of the system. Those variations significantly affect the propagation of guided waves on the structure \cite{dutta2022time}. More specifically, guided waves are sensitive to the loading conditions, and environmental effects, such as temperature, humidity, and moving constraints of the structure \cite{raghavan2008effects}. When dealing with structures made of materials with temperature-dependent properties, the environmental effects in particular may influence the behavior of the structure. Moreover, there is often uncertainty in the material properties of the different structural components related to the manufacturing process of the same material. This fact leads to identical structural components having slightly different material properties \cite{ahmed2019uncertainty}. These uncertainties have an important effect on the prognostic capability of the SHM system and may lead to misidentified damage, false alarms, or missed fault detection \cite{cross2022physics}. In this direction, uncertainty quantification (UQ) is often applied in real-world structures, aiming to improve the reliability of the prediction for probabilistic decision-making by quantifying the sources of variability and uncertainty in the model inputs \cite{ahmed2019uncertainty}.  

Along these lines, computational mechanics has proven so far to be a unique tool for developing high-fidelity models (HFM) that capture with great accuracy the response of the actual structure. When integrated into an SHM system, numerical simulation can improve diagnostic and prognostic performances \cite{farrar2007damage}. In the frame of structural mechanics, numerical methods, such as the finite element method (FEM) \cite{polyzos2021numerical} and the boundary element method \cite{gortsas2017numerical, rodopoulos2020nonlinear, drakoulasnumerical}, are widely utilized to simulate the GWP \cite{bellam2021parametric}. To accomplish that, a high-quality mesh grid and a fine temporal discretization are essential to obtain accurate predictions, especially in the high-frequency range \cite{willberg2012comparison, borate2020data}. Therefore, the computational cost is increased, and the need for high-performance computing is often essential to simulate large-scale, transient problems.
Consequently, for the realization of the DT concept, surrogate modeling has gained attention, aiming to reduce the simulation cost of HFMs while providing accurate numerical results \cite{drakoulas2023fastsvd}. Reduced order modeling (ROM), as part of surrogate modeling, is commonly used to lessen the computational burden of large-scale, transient structural simulations by predicting the solutions in a limited amount of time \cite{rasheed2019digital}. Non-intrusive ROM (NIROM) based on proper orthogonal decomposition (POD) and deep learning (DL) methods turns out to be an efficient technique to significantly decrease the simulation time while preserving the dominant effects in the structural response of the numerical models \cite{simpson2021machine, nikolopoulos2022machine}. 

To deploy a NIROM using DL methods, a high-fidelity numerical dataset is essential, the size of which varies depending on various factors, such as the ranges of the parameter values, the time period of the simulation, and the (non)linear behavior of the physics-based problem \cite{drakoulasHSTAM}. Therefore, an increased number of parameters, related to the boundary and initial conditions, material properties, and geometrical characteristics, is often essential to defining the HFM and thus makes it challenging to deploy ROMs with a limited amount of numerical data \cite{wang2021novel}. To tackle this, various adaptive sampling strategies have been proposed in the literature, aiming to improve the training of the ROM by identifying appropriate parameters for which the HFM results are obtained. The sampling strategies are divided into one-shot and sequential methods \cite{karcher2022adaptive}. In the former group, amongst others, the Monte Carlo, the Halton sequence, and the Latin hypercube sampling (LHS) \cite{stein1987large} are utilized to predefine the parameter sets for the ROM training. On the contrary, sequential adaptive sampling methods choose the parameter values iteratively, providing useful insight into ROM training. Greedy algorithms are often utilized for sequential sampling, by selecting the parameters that maximize an estimated error \cite{chellappa2021training, chen2018greedy, dutta2021greedy}. Sequential sampling methods, based on Bayesian optimization (BO) and active learning combined with Gaussian process regression (GPR) \cite{polyzos2021ensemble, oikonomou2022physics, zhuang2022active}, have proven to be an alternative active sampling method in a probabilistic manner \cite{kast2020non}. 

In the last few decades, there has been intensive research on the integration of machine learning (ML) methods and physics-based simulations for SMH applications, targeting the development of comprehensive DTs \cite{colombera2021generative, malekloo2022machine}. In the field of surrogate modeling, various hybrid systems have been proposed, aiming to combine ROMs with data-driven approaches for SHM applications focusing on damage detection, localization, and characterization \cite{rosafalco2021online, torzoni2020combined, torzoni2022shm, torzoni2022structural, muralidhar2022damage, torzoni2023deep, tatsis2022hierarchical}. In the context of this work, we are targeting ML-based ROM frameworks in the SHM field, utilizing ultrasonic-guided waves, to inspect the structure and provide probabilistic estimations of the effects on the quantities of interest caused by perturbations in the structural integrity \cite{drakoulasmachine}. Closest to our approach are the methodologies presented in \cite{borate2020data, muralidhar2022damage, bellam2021parametric, sharma2023wave, mikhaylenko2022numerical, bigoni2022predictive, sieber2023guided}. In \cite{borate2020data, mikhaylenko2022numerical, bellam2021parametric}, the authors proposed data-driven SHM approaches for virtual sensing using gappy POD using GWP responses, focusing on damage detection and identification. In \cite{muralidhar2022damage}, the authors presented a POD-based ROM for damage localization and characterization using GWPs and Bayesian modeling. In another work \cite{sharma2023wave}, a DL-based surrogate model is introduced to localize and identify the depth of the crack using guided waves. In \cite{bigoni2022predictive}, a ROM based on the reduced basis (RB) method and Galerkin projection is proposed for optimal sensor placement while providing the state of the damage in the structures. In a recent work \cite{sieber2023guided}, the authors introduced a ROM for GWP deployed in the frequency domain to decrease the training data, targeting to localize damages in a composite panel.

In this study, an innovative ML-based ROM framework is presented, to simulate large-scale guided wave propagation problems in a limited amount of time, while quantifying the impact of the uncertainties in the material properties of thin, lightweight structures. The proposed framework, referred to as \textit{BO-ML-ROM}, extends the \textit{FastSVD-ML-ROM} \cite{drakoulas2023fastsvd}, by introducing a GPR-based, sequential, adaptive sampling BO methodology \cite{garnett2023bayesian}. A predefined error is selected for the BO to control the accuracy and the convergence of the algorithm, acting also as a stopping criterion during the simulation of the HMFs. Compared to classical one-shot sampling techniques, such as LHS, our approach increases the accuracy of the predictions, while reducing the time required for the ROM preparation, including the data-gathering process and the training of the DL models. More specifically, in the developed framework, the GP-based BO is merged with a singular value decomposition (SVD) update methodology \cite{drakoulas2023fastsvd}, to compute a linear reduced basis using the numerical results. Furthermore, the \textit{BO-ML-ROM} includes, \textit{i)} a convolutional autoencoder (CAE) for non-linear dimensionality reduction; \textit{ii)} a feed-forward neural network (FFNN) to map the input parameters to the latent spaces extracted through CAEs, and \textit{iii)} a long-short-term memory (LSTM) network to predict the dynamics of the parametric solutions. 

For the numerical simulation of the GWP problem, the FEM is utilized. The developed methodology is applied to a UQ problem, related to the GWP in an aluminum, rectangular plate. In particular, the impact on the GWP, caused by the variations in the material properties of the plate, is examined. These variations account both for environmental and manufacturing procedures. On top of that, a global, variance-based sensitivity analysis (SA) using Sobol' indices \cite{nossent2011sobol} is performed to estimate the contribution of each source of uncertainty to the predicted results. The predicted results, demonstrate the efficiency of the \textit{BO-ML-ROM} compared to the LHS, both in terms of speed-up of the data gathering process, and accuracy. Furthermore, the applicability of the developed ROM for the simulation of the GWP in a decreased time is presented, and its utility for the UQ and SA, which are prohibitive to perform with the utilized FEM solver due to the increased computational cost, is revealed.

The current work proposes several novelties, and to the best of our knowledge, there is no other ROM framework in the SHM community with the following features:
\begin{enumerate}[nolistsep]
  \setlength{\itemsep}{0.05pt}
  \setlength{\parskip}{0.03pt}
  \setlength{\parsep}{0.03pt}
\item A ML-based ROM is utilized to simulate large-scale GWP and predict the acceleration field in a limited amount of time. This is achieved through a hyper-reduction technique using both linear and non-linear methods, as well as LSTMs for the temporal evolution of the GWP. 
\item A BO framework based on GPR is used to optimally choose the parameter vectors for the formulation of the high-fidelity data. A predefined error is employed in the BO framework, acting as a stopping criterion that controls the efficiency of the utilized active sampling methodology. 
\item The \textit{BO-ML-ROM} is used to address the forward UQ for a GWP problem and examine the effects of the variations in the material properties caused by both manufacturing and temperature effects in the acceleration fields. 
\item A variance-based SA is performed to study the amount of uncertainty apportioned to the different sources of variation in the acceleration fields. 
\end{enumerate}

The structure of this paper is as follows: In Section \ref{GWP}, the dynamic behavior of a linear elastic material and the solution of the HFM are presented. In Section \ref{BO}, the BO algorithm for adaptive parameter sampling is analyzed, and the integration with the SVD update algorithm is presented, in detail. The structure of the \textit{BO-ML-ROM} based on the integration of the BO algorithm and the DL models are described in Section  \ref{FastSVDMLROM}. Furthermore, the forward UQ and SA procedures implemented in this work, are described in Section \ref{ROMUQ}. In Section \ref{application}, the \textit{BO-ML-ROM} is applied to predict the GWP and estimate the impact of the material uncertainties in a thin, rectangular aluminum plate. Finally, in Section \ref{Conclusions}, a discussion of the presented results is performed, and some potential future works related to the coupling of SHM and ROM, are listed.

\section{Problem Formulation} \label{GWP}
In this section, the equations related to the dynamical behavior of a linear elastic material are briefly presented. For the sake of simplicity, the equations are provided for a 3D solid. Regarding thin structures, such as plates and shells, it is often preferable to use the appropriate engineering assumptions, which lead to various reduced theories \cite{timoshenko1959theory, reddy2003mechanics}.

Consider a linear elastic solid occupying a volume $\Omega \subset \mathbb{R}^d$, where $d$ is the number of spatial dimensions. The balance of momentum that describes the dynamical behavior of the solid can be expressed as follows, 
\begin{equation} \label{general}
    \nabla \cdot \bm{\sigma}(\bm{x},t) - \rho(\bm{x}) \ddot{\bm{u}}(\bm{x},t) + \bm{f}(\bm{x},t) = 0, \:\:\: \bm{x} \in \Omega,  \:\:\: t \in (0,t_d]
\end{equation}
where $\bm{x}$ is a material point, $\bm{\sigma}$ is the Cauchy stress tensor, $\bm{f}$ are the volume forces, $\rho$ is the density, $\bm{u}$ denotes the displacement field, $\ddot{\bm{u}}$ represents the acceleration field, and $(0,t_d]$ is the temporal domain. The constitutive behavior of the material is described by the following equations,
\begin{equation} \label{Equation Material}
    \bm{\sigma}(\bm{x},t)={\mathbb{C}}(\bm{x}):\bm{\epsilon}(\bm{x},t) \: \Leftrightarrow \: \bm{\epsilon}(\bm{x},t)={\mathbb{S}}(\bm{x}):\bm{\sigma}(\bm{x},t)
\end{equation}
where $\mathbb{C}$ is a fourth-order stiffness tensor and its structure depends on the directional material properties, $\mathbb{S} = \mathbb{C}^{-1}$ is the compliance tensor, and $\bm{\epsilon}$ is the small strain tensor given as,
\begin{equation}
    \bm{\epsilon}(\bm{x},t) = \frac{1}{2} \left[ \nabla \bm{u}(\bm{x},t)+\nabla^T \bm{u}(\bm{x},t)\right ]
\end{equation}
The boundary conditions related to Eq. \ref{general} are the following,
\begin{subequations}\label{eq:4}
\begin{align}
  \bm{u}(\bm{x},t) = \bar{\bm{u}}(\bm{x},t),  \:\: \bm{x} \in \partial{\Omega}_{\bm{u}}  \:\:\:  \\
  \bm{t}(\bm{x},t) = \bm{n}(\bm{x}) \cdot \bm{\sigma}(\bm{x},t) = \bar{\bm{t}}(\bm{x}), \:\:  \bm{x} \in \partial{\Omega}_{\bm{t}}
\end{align}
\end{subequations}
where $\bm{n}(\bm{x})$ is the unit normal vector to the point $\bm{x}$ and $\bm{t}$ is the traction vector. The initial conditions are,
\begin{equation} \label{Compliance mat}
         \bm{u}(\bm{x},0) = \bm{u}_0(\bm{x}),  \:\:\:
         \dot{\bm{u}}(\bm{x},0) = \dot{\bm{u}}_0(\bm{x}), \:\: \bm{x} \in \Omega
\end{equation}
Using Voigt notation \cite{sadd2009elasticity}, the relation between the strains and the stresses (Eq. \ref{Equation Material}) can be written in the following form:
\begin{equation}
\begin{bmatrix} 
\mathbf{\epsilon}_{x} \\ \mathbf{\epsilon}_{y} \\  \mathbf{\epsilon}_{z} \\ 2\mathbf{\epsilon}_{xy} \\ 2\mathbf{\epsilon}_{yz} \\ 2\mathbf{\epsilon}_{zx}
\end{bmatrix}
=
\begin{bmatrix} 
{S}_{11} & {S}_{12} & \cdot & \cdot & \cdot & {S}_{16}\\
{S}_{21} & \cdot           & \cdot & \cdot & \cdot & \cdot\\
\cdot           & \cdot           & \cdot & \cdot & \cdot & \cdot\\
\cdot           & \cdot           & \cdot & \cdot & \cdot & \cdot\\
\cdot           & \cdot           & \cdot & \cdot & \cdot & \cdot\\
{S}_{61} & \cdot           & \cdot & \cdot & \cdot & {S}_{66}\\
\end{bmatrix}
\begin{bmatrix} \label{matrix_} 
\mathbf{\sigma}_{x} \\ \mathbf{\sigma}_{y} \\  \mathbf{\sigma}_{z} \\ \mathbf{\tau}_{xy} \\ \mathbf{\tau}_{yz} \\ \mathbf{\tau}_{zx}
\end{bmatrix}
\end{equation}
where $S_{11}, S_{12}, \ldots, S_{66}$ are the terms of the compliance matrix.

The weak formulation corresponds to the problem described by Eqs. (\ref{general}), (\ref{eq:4}), and (\ref{Compliance mat}), can be expressed as, 
\begin{equation} \label{eq:7}
   \int_{\Omega} \rho(\bm{x})\bm{w}(\bm{x}) \cdot \ddot{\bm{u}}(\bm{x},t) \:  d\Omega +      \int_{\Omega} \nabla \bm{w}(\bm{x}): \bm{\sigma}(\bm{x},t) \: d\Omega  = \int_{S} \bm{w}(\bm{x}) \cdot \bm{t}(\bm{x},t) \: dS +      \int_{\Omega} \bm{w}(\bm{x})\cdot \bm{f}(\bm{x},t) \: d\Omega 
\end{equation}
where $\bm{w}(\bm{x})$ denotes the test function. Both $\bm{w}(\bm{x})$ and $\bm{u}(\bm{x})$ belong to an appropriate Sobolev space, containing functions of finite energy \cite{hughes2012finite}.

Utilizing the discretization of the domain $\Omega$ with finite elements, Eq. \ref{eq:7} can be expressed as follows,
\begin{equation} \label{ref_PDE}
    \bm{M}(\bm{\theta}) \ddot{\bm{u}_h}(t; \bm{\theta}) + \bm{K}(\bm{\theta}) \bm{u}_h(t; \bm{\theta}) = \bm{f}(t),  \  \ t\in (0,T]
\end{equation}
with $\bm{M}$ representing the global mass matrix, $\bm{K}$ denotes the global stiffness matrix, and $\bm{\theta}$ denoting the parameter vector, sampled from a parameter space $\mathcal{P}$ $\subset$ ${\mathbb{R}}^{\xi}$, related to the variations in the material properties (density and stiffness). The dimension $\xi$ of the vector $\bm{\theta}$ depends on the subset of the chosen material parameters to study the parametric solution of Eq. (\ref{ref_PDE}). The discretization parameter, $h>$0, is related to the element size. The parameterized discrete FOM solution, is denoted as $\bm{u}_h(t;\bm{\theta})$: $\mathbb{R}^+ \times \mathbb{R}^{\xi}$ $\rightarrow$ $\mathbb{R}^{N_h}$, and $N_h$ is the dimension of $\bm{u}_h$.  

In the context of this work, Gaussian distributions are assigned to each feature of the parameter vector $\bm{\theta}$, part of the space $\mathcal{P}$. Three different sets are defined to deploy the ML-based ROM platform for the UQ based on the BO; ${\theta}^{tr}$ = $\{\bm{\theta}_1^{tr},\ldots,\bm{\theta}_m^{tr}\}$, that corresponds to the $m$ training parameter vectors selected using the BO, ${\theta}^{te}$ = $\{\bm{\theta}_1^{te}, \ldots,\bm{\theta}_n^{te}\}$  that contains the  $n$ testing vectors, sampled randomly from the parameter space to test the ROM platform, and ${\theta}^{uq}$=$\{\bm{\theta}_1^{uq}, \ldots,\bm{\theta}_r^{uq}\}$, that corresponds to the $r$ vectors sampled from the multivariate Gaussian distribution of the parameter vectors to estimate the UQ. It is worth mentioning that, $\theta^{tr}\cup\theta^{te}\cup \theta^{uq}\subset{\mathcal{P}}$ and $\theta^{te} \cup \theta^{tr} \ll \theta^{uq} $, since each solution for the UQ is computed with the trained ROM framework and can be obtained in a limited amount of time. 

The dataset matrix $\bm{S}_h (\bm{\theta}_j^{tr}) \in \mathbb{R}^{N_h \times N_t}$ composed of the FOM solutions, for each training parameter vector $\bm{\theta}_{j}^{tr}$ with $j=1,\ldots,m$, and for the discrete-time values {$t_1, \ldots, t_{N_t}$} with $t_i \in (0, t_d]$, is the following,
\begin{equation} \label{matrix_data}
    \bm{S}_h (\bm{\theta}_j^{tr}) = [\; \bm{u}_h(t_i;\bm{\theta}^{tr}_j) \;| \ldots| \; \bm{u}_h(t_{N_t};\bm{\theta}^{tr}_j) \;]
\end{equation}

\section{Bayesian optimization (BO) for adaptive parameter sampling} \label{BO}
In this section, the methodology for the adaptive parameter sampling based on BO is presented. The essential components of the BO framework, such as the acquisition functions as well as the kernels of the GPR models, are described. Furthermore, the algorithm for the integration of BO with the iterative computation of the reduced linear basis, aiming to identify the training parameters for which the HFMs are computed, is outlined.

\subsection{Reduced linear basis computation based on SVD update}
An essential component for the implementation of the \textit{BO-ML-ROM}, is the application of an SVD update algorithm \cite{drakoulas2023fastsvd}, to compute a linearly reduced basis. The SVD update used in this work generates a single reduced linear basis for the entire snapshot matrix iteratively instead of performing the reduction directly. The purpose of this approach is to compress the snapshot matrix incrementally during the data-gathering process by partitioning each HFM solution $\bm{u}_h(t_i;\bm{\theta}^{tr}_j)$ in several batches, based on the computer resources. Consequently, as the snapshot matrix is assembled, it is also gradually compressed. The computed basis remains the same during the ROM deployment in the online phase, i.e., it is not parameter-dependent. Naturally, the application of the initial SVD contributes to the total approximation error of the ROM and needs to be small enough, since the procedures that follow increase the initial error. 

The aforementioned SVD update methodology is integrated with the BO framework, aiming to incrementally select the training parameter vectors $\bm{\theta}_j^{tr}$ with $j=1, \ldots, m$, in an adaptive manner, and train the \textit{BO-ML-ROM}. During the incremental sampling of the parameters $\bm{\theta}^{tr}$, the basis matrix is updated. In the following, we denote as $\bm{U}_j$, the basis matrix, which has been formed up to the $j$-th chosen parameter. To accomplish the adaptive sampling, the l2-norm error estimator $\varepsilon^{tr}(\bm{\theta}^{tr})$ that corresponds to $\bm{\theta}^{tr}$, is calculated in each step of the procedure for all the chosen parameters up to the $j$-th step based on the projection of each HFM solution $\bm{S}_h(\bm{\theta}^{tr})$ to the linear reduced basis stored in the matrix $\bm{U}_j$, as follows
\begin{equation} \label{l2-norm}
        {\varepsilon^{tr}_\lambda}={\varepsilon^{tr}}(\bm{\theta}^{tr}_\lambda) = 
        \frac{||{\bm{S}_h(\bm{\theta}^{tr}_\lambda)- {\tilde{\bm{S}_h}(\bm{\theta}^{tr}_\lambda)}}||_F } {||\bm{S}_h(\bm{\theta}^{tr}_\lambda)||_F}, \: \: \lambda=1,\ldots,j
    \end{equation}
where $\lambda$ is a counter of the training parameters until the $j$-th step, $||{\bm{S}}(\bm{\theta}^{tr}_\lambda)||_F=\sqrt{\sum_{a=1}^{N_h}\sum_{b=1}^{N_t} {S^2_{ab}(\bm{\theta}_\lambda^{tr})}}$, and $\bm{\tilde{S}}_h(\bm{\theta}^{tr}_\lambda)$ is the reconstructed solution given as,
\begin{equation}\label{eq14}
    \Tilde{\bm{S}}_h(\bm{\theta}_\lambda^{tr}) = {\bm{U}_{j}} \ \bm{U}_{j}^T \bm{S}_h(\bm{\theta}_\lambda^{tr})
\end{equation} 
The sets $\{ (\bm{\theta}^{tr}_\lambda, {\varepsilon^{tr}_\lambda} ) \}_{\lambda=1}^j$ are used in each step $j$, of the BO methodology for the formulation of the labeled dataset. The truncation in the SVD algorithm is computed with respect to a predefined error $\varepsilon_{svd}$, which is a parameter of the algorithm.

\subsection{Bayesian Optimization}
In the context of this work, a BO framework is implemented to calculate the most informative parameters $\bm{\theta}^{tr}$ for which the HFM solutions $\Tilde{\bm{S}}_h(\bm{\theta}^{tr})$ are computed, and the basis matrix $\bm{U}$ is formed. BO is a class of probabilistic optimization algorithms that aim to identify the global optimum of \textit{black-box} functions incrementally using a limited number of function evaluations by incorporating both exploration and exploitation \cite{frazier2018tutorial}. The BO methodology utilizes a statistical surrogate model $f(\cdot)$, to map the input features to the function outputs. The BO framework is used to enrich the training dataset and the basis $\bm{U}$ incrementally through an adaptive sampling procedure with a predefined error, $\varepsilon_{tol}$, acting as a stopping criterion. The algorithm is terminated once the mean value of the computed l2-norm errors $\varepsilon_j^s$ (Eq. \ref{error}) in the $j$-th iteration for the testing parameters is lower than the predefined error $\varepsilon_{tol}$.
\begin{equation} \label{error}
    \varepsilon_j^s = \frac{1}{n} \sum^{n}_{a=1} {\frac{||{\bm{S}_h(\bm{\theta}^{te}_a)- \bm{U}_{j}} \ \bm{U}_{j}^T\bm{S}_h(\bm{\theta}^{te}_a)||_F } {||\bm{S}_h(\bm{\theta}^{te}_a)||_F}}
\end{equation}

In this work, the features ${\theta}^{tr}_{\upsilon}$, with $\upsilon=1, \ldots, \xi$, where $\xi$ is the total number of features that compose the parameter vector $\bm{\theta}^{tr}$, are considered Gaussian random variables. Consequently, ${\theta}^{tr}_{ \upsilon} \sim \mathcal{N}(m_\upsilon, s_\upsilon^2)$, where $m_\upsilon$ and $s_\upsilon$ are the mean and standard deviation of the feature ${\theta}^{tr}_{\upsilon}$. This is a common assumption for the material properties of the structural components \cite{ahmed2019uncertainty}. The optimization procedure requires a minimum and maximum value of the parameters ${\theta}^{tr}_{\upsilon}$. To capture 99.99 \% of the observations, for each feature, we define $\theta^{tr}_{\upsilon} \in [m_\upsilon - 4s_\upsilon, m_\upsilon + 4s_\upsilon]$. 

Before the initial application of the BO procedure, an initial dataset of $\tau$ parameters $\bm{\theta}^{tr}$, with, $\tau\ll m$ is randomly sampled using the LHS method. The size of the initial dataset $\tau$ is chosen in an ad-hoc manner. For these parameters, the HFM solutions are computed, and the SVD update algorithm is used to calculate the reduced linear basis $\bm{U}_\tau$. Next, the generated l2-norm errors are calculated (Eq. \ref{l2-norm}), and a labeled dataset is formed, $\mathcal{D}_\tau:=\{ (\bm{\theta}_\alpha^{tr}, \varepsilon_\alpha^{tr}) \}_{\alpha=1}^{\tau}$. 

The BO algorithm utilizes an acquisition function, $\alpha(\cdot)$ to guide the exploration and exploitation of the model $f$ in the search space $\mathcal{P}$, which constitutes also the parameter space, that contains all the feasible values of the parameter vector $\bm{\theta}^{tr}$ being in agreement with the bounds of the material properties \cite{wilson2018maximizing}. The main objective of the BO is to systematically search the space $\mathcal{P}$, for the sequential acquisition of the global maximum value as follows, 

\begin{equation} \label{Eq:12}
    \bm{\theta}_{\tau+1}^{tr} = \underset{\bm{\theta}^{tr}\in\mathcal{P}}{\arg\max} \: \alpha(\bm{\theta}^{tr}|\mathcal{D}_{\tau}), \:\:\: given \:\: p(f(\bm{\theta}^{tr})|\mathcal{D}_\tau)
\end{equation}

The BO procedure is applied for the choice of the remaining $m -\tau +1$ parameters. In each of the following steps, the labeled dataset is updated using the information related to all the previously chosen parameters. Therefore, the important components for the implementation of the BO framework are, \textit{i)} the design of the statistical model $f(\cdot)$ aiming to obtain the posterior probability $p(f(\bm{\theta}^{tr})|\mathcal{D}_\tau)$ given the dataset $\mathcal{D}_{\tau}$, and, \textit{ii)} the choice of the AF $\alpha(\cdot)$, that aims to identify the next parameter vector $\bm{\theta}_{\tau+1}^{tr}$, by balancing the exploration and exploitation in the statistical model $f$.

\subsubsection{Gaussian process regression (GPR)} \label{Ch: GPR}
The BO algorithm requires a stochastic-based surrogate model $f$, to map each input parameter vector, $\bm{\theta}^{tr}$, to the respective l2-norm error, $\varepsilon^{tr}$ accounting for the uncertainties in the predictions. Among the various probabilistic ML-based models such as the Bayesian neural networks \cite{jospin2022hands} and polynomial chaos expansion \cite{sudret2008global}, Gaussian processes are widely utilized within the context of BO frameworks \cite{10095008}. GPR is a convenient class of non-parametric regression models utilizing the properties of the multivariate normal distribution for supervised learning problems \cite{garnett2023bayesian}. 

The GPR constitutes a surrogate model for the regression function $f$, that maps the input feature vectors $\bm{\theta}^{tr}$ to the corresponding $\varepsilon^{tr}$ (Eq. \ref{l2-norm}). It assumes that the range of the function $f$, with, $f: \mathcal{R}^{\xi} \rightarrow \mathcal{R}$ is composed of Gaussian random variables. Consequently, a GP prior is imposed on  $f$ as $f \sim \mathcal{GP}(0, \kappa(\bm{\theta}, \bm{\theta}^{'}))$, where the function $\kappa(\cdot, \cdot)$, also called covariance function, with $\kappa: \mathcal{R}^{\xi} \times \mathcal{R}^{\xi} \rightarrow \mathcal{R}$, is a positive-definite kernel that measures the correlation between the pair of parameter vectors $(\bm{\theta}, \bm{\theta}^{'})$ \cite{10095008}. In the frame of this work, the squared exponential radial basis (rbf) and the mattern (mtk) functions \cite{frazier2018tutorial} are used to compute the covariance terms. The rbf function is defined using the following formula, 
\begin{equation}
   \kappa_{rbf}(\bm{\theta}, \bm{\theta}^{'}) = \sigma_r^2 exp(- \frac{||\bm{\theta}-\bm{\theta}^{'}||^2}{2l_r})
\end{equation}
where $l_r$ is the length scale parameter \cite{buhmann2000radial}, $\sigma_{r}$ is the variance parameter, and $||\bm{\theta}-\bm{\theta}^{'}||$ denoted the Euclidean distance between the two parameter vectors $\bm{\theta}, \bm{\theta}^{'}$, while the mtk function is given by the following expression,
\begin{equation}
   \kappa_{mtk}(\bm{\theta}, \bm{\theta}^{'}) = \frac{2^{1-\nu_m}}{\Gamma(\nu_m)} \left (  \frac{\sqrt{2\nu_m}||\bm{\theta}-\bm{\theta}^{'}||}{l_m}\right )^{\nu_m} K_{\nu_m} \left ( \frac{ \sqrt{2\nu_m} ||\bm{\theta}-\bm{\theta}^{'}||}{l_m} \right ) 
\end{equation}
where $K_{\nu_m}(\cdot)$ is the modified Bessel function of the second kind, ${\nu}_m$ is a positive parameter, and $\Gamma(\cdot)$ is the Gamma function \cite{rasmussen2006gaussian}. In this work, the mtk kernel is used with $\nu_m=1.5$. Moreover, a combination of the rbf and mtk functions, $k_{rbf}*k_{mtk}$ is used to define an additional kernel aiming to further evaluate the performance of the GPR. 

The initial dataset $\mathcal{D}_\tau$ that includes the parameter vectors $\bm{\Theta}_\tau =[\bm{\theta}_1^{tr}, \ldots, \bm{\theta}_\tau^{tr}]^{\top}$, is used to compute the prior function values, $\bm{f}_\tau =[f(\bm{\theta}^{tr}_1), \ldots, f(\bm{\theta}^{tr}_\tau)]^{\top}$ that are Gaussian distributed, with $ p(\bm{f}_\tau|\bm{\Theta}_\tau)=\mathcal{N} (\bm{f}_\tau;\bm{0}_{\tau}, \bm{K}_\tau)$, where $\bm{K}_\tau$ represents a $\tau  \times \tau$ covariance matrix, and an element of the matrix, corresponding to the index pair $(a, b)$, with $a,b = 1, \ldots, \tau$, is equal to $[\bm{K}_{\tau}]_{a, b}=\kappa(\bm{\theta}_{a}^{tr}, \bm{\theta}_{b}^{tr})$.  
The main objective of the regression task is to compute the distribution of the function value $f(\bm{\theta}^{tr})$, for a new generic input parameter $\bm{\theta}^{tr}$. The joint distribution of $f(\bm{\theta}^{tr})$ and, $\bm{f}_\tau$ can be represented as follows, 
\begin{equation} \label{eq:16}
    \left[
    \begin{array}{cc}
         {\bm{f}_\tau} \\ f(\bm{\theta}^{tr})
    \end{array}
    \right]\sim \mathcal{N} 
    \left[
    \begin{array}{cc} 
    \left[
	   \begin{array}{cc}
	         0 \\
	         0 
	    \end{array}
     \right],
	    & 
	    \left[
	    \begin{array}{cc}
	        \bm{K}_\tau & \bm{k}_{\tau}(\bm{\theta}^{tr})) \\
	         \bm{k}_{\tau}^T(\bm{\theta}^{tr}) & {\kappa}(\bm{\theta}^{tr},\bm{\theta}^{tr}) 
	    \end{array}
	    \right]
	\end{array}
	\right]
\end{equation}
where $\bm{k}_\tau(\bm{\theta}^{tr}):=[\kappa(\bm{\theta}_{1}^{tr},\bm{\theta}^{tr}), \ldots, \kappa(\bm{\theta}_{\tau}^{tr},\bm{\theta}^{tr})]^{\top}$.
Using the Bayesian approach for the labeled dataset $\mathcal{D}_\tau$ and the pdf of Eq. \ref{eq:16}, the posterior of $f(\bm{\theta}^{tr})$, can be obtained as \cite{rasmussen2004gaussian},
\begin{equation}
    p(f(\bm{\theta}^{tr})|\mathcal{D}_\tau) = \mathcal{N}(f(\bm{\theta}^{tr}); \mu_\tau(\bm{\theta}^{tr}), \sigma^2_\tau(\bm{\theta}^{tr}))
\end{equation}
where the mean $\mu_\tau$, and variance $\sigma_\tau^2$ can be computed as follows,
\begin{equation}
    \mu_\tau(\bm{\theta}^{tr}) = k_\tau^T(\bm{\theta}^{tr})(\bm{K}_\tau)^{-1}\bm{f}_\tau
\end{equation}
\begin{equation}
    \sigma_\tau^2(\bm{\theta}^{tr}) = \kappa(\bm{\theta}^{tr},\bm{\theta}^{tr})-\bm{k}^{T}(\bm{\theta}^{tr})(\bm{K}_\tau)^{-1}\bm{k}_\tau(\bm{\theta}^{tr})
\end{equation}
The hyperparameters of the GPR model, are obtained through the maximum likelihood estimation theorem using the Scikit-learn library \cite{sklearn_api, rasmussen2004gaussian}.
\begin{comment}
The hyperparameters of the GPR model, are obtained through the maximum likelihood estimation (MLE) theorem using the sklearn library \cite{sklearn_api}, using the following formula,
\begin{equation} \label{MLE}
    log \pi(\bm{y}|\bm{X}, \bm{\theta}) = - \frac{1}{2}(\bm{y}^T(K(\bm(X),\bm(X))+\sigma^2_y\bm{I})^{-1}\bm{y} + log det(K(\bm{X},\bm{X}+\sigma_y^2\bm{I})) + N_{data}log2\pi
\end{equation}
Therefore, we obtain the hyperparameters $\theta=(\sigma_y, \sigma_f, l)$ of the RBF kernel, we utilize MLE (Eq. \ref{MLE}), on the training dataset.  
\end{comment}

\subsubsection{Acquisition function (AF)} \label{Ch: AF}
Moving forward, the statistical surrogate model $f$ is utilized, to obtain the next parameter vector $\bm{\theta}^{tr}_{\tau+1}$ through Eq. \ref{Eq:12}, using an acquisition function $\alpha(\cdot)$. The function $\alpha$ is designed to balance exploration and exploitation \cite{noe2018new}. In particular, two AFs are tested, namely: the probability of improvement (PI) and the expected improvement (EI) functions \cite{snoek2012practical}. In particular, the PI and EI functions calculate the expectation of the improvement on $f$, with respect to the predictive distribution of the probabilistic surrogate model.

The PI considers the probability of improving the current best estimate. Using the GPR as a regression model, PI can be defined as, 
\begin{equation}
    \alpha_{PI} (\bm{\theta}_{\tau+1}^{tr};\mathcal{D}_\tau) = \Phi \left( \frac{\Delta (\bm{\theta}_j^{tr})}{\sigma(\bm{\theta}_j^{tr})}  \right)
\end{equation}
with $\Delta_\tau(\bm{\theta}_j^{tr}) = \mu_\tau(\bm{\theta}_j^{tr}) - \hat{f}_\tau^{max} -\xi_\alpha$, and $\Phi(\cdot)$ denotes the normal cumulative distribution function; $\xi_\alpha$ is a parameter which modulates the balance between exploration and exploitation, $ \hat{f}_\tau^{max}$ is the maximum obtained value of the surrogate model $f$ related to the dataset $\mathcal{D}_\tau$, given by $\hat{f}_\tau^{max}=max(\varepsilon_1^{tr}, \cdots, \varepsilon_\tau^{tr} )$ \cite{archetti2019bayesian}.

However, the PI function aims to select points without accounting for the magnitude of improvement. To overcome this issue, the EI acquisition function has been proposed \cite{archetti2019bayesian}, leading to a better balance between exploration and exploitation \cite{polyzos2023bayesian}. Therefore, the EI can be computed as,

\begin{equation}
    \alpha_{EI} (\bm{\theta}_{\tau+1}^{tr};\mathcal{D}_\tau) =  \sigma_\tau(\bm{\theta}_j^{tr})\phi \left( \frac{\Delta_\tau(\bm{\theta}_j^{tr})}{\sigma_\tau(\bm{\theta}_j^{tr})} \right) + \Delta_\tau(\bm{\theta}_j^{tr}) \Phi \left (\frac{\Delta_\tau(\bm{\theta}_j^{tr})}{\sigma_\tau({\bm{\theta}_j^{tr}})} \right)
\end{equation}
with $\phi(\cdot)$ denoting the Gaussian pdf.  

\subsection{Algorithm for sequential parameter sampling through BO and linear reduced basis computation}
In this subsection, the entire algorithm of the BO for optimum parameter sampling, including the integration of the GPR model, the acquisition function, and the incremental linear reduced basis formulation, is presented in Algorithm \ref{algorithm}. 

\begin{algorithm} 
\caption{BO and linear reduced basis computation}\label{algorithm}
      \hspace*{\algorithmicindent} \textbf{Input} : Number of testing parameters $n$, number of initial parameters $\tau$, parameter space $\mathcal{P}$, truncation error $\varepsilon_{svd}$, stopping error $\epsilon_{tol}$, acquisition function $\alpha(\cdot)$, kernel $k(\cdot, \cdot)$.  \\
       \hspace*{\algorithmicindent} \textbf{Output} : Training parameters $\theta^{tr}$, testing parameters $\theta^{te}$, basis matrix $\bm{U}$, training HFM solutions $\bm{S}_h(\bm{\theta}_{j}^{tr})$ with $j=1,\ldots,m$, testing HFM solutions $\bm{S}_h(\bm{\theta}_{j}^{te})$ with $j=1,\ldots,n$.
        \begin{algorithmic}[1]
    \State  Sample $n$ testing and $\tau$ initial parameter vectors from the parameter space $\mathcal{P}$ using LHS. 
    \State Compute the $(\tau+n)$ HFM datasets $\bm{S}_h$ (Eq. \ref{matrix_data}), using the FEM.
    \State Utilizing the SVD update algorithm, compute the basis $\bm{U}_\tau$ using the dataset for the parameter vectors $\bm{\theta}^{tr}_a$ with $a=1,\ldots,\tau$.
    \State Compute the errors $\varepsilon_\lambda^{tr}$ (Eq. \ref{l2-norm}), and construct the labeled dataset $\mathcal{D}_\tau:=\{(\bm{\theta}_\lambda^{tr}, \varepsilon_\lambda^{tr)}\}_{_\lambda=1}^{\tau}$.
    \State Set $j=\tau$ and $\varepsilon_{tol}=\infty$
    \While {$\varepsilon^s_j$ $\geq$ $\varepsilon_{tol}$ }
           \State Train the GPR model, with the labeled dataset $\mathcal{D}_j$, and obtain the surrogate model $f$. 
           \State Using the model $f$ and function $\alpha(\cdot)$, obtain the parameter vector $\bm{\theta}_{j+1}^{tr}$ (Eq. \ref{Eq:12}). 
           \State Compute the HFM dataset $\bm{S}_h(\bm{\theta}_{j+1}^{tr})$.
           \State Using SVD update, compute the basis matrix $\bm{U}_{j+1}$.
           \State Recompute the errors $\varepsilon_q$ with $q=1,\ldots, j+1$ and form the new labeled dataset $\mathcal{D}_{j+1}$.
           \State Compute the l2-norm $\varepsilon^{s}_{j+1}$, using the testing HFMs and the basis $\bm{U}_{j+1}$ (Eq. \ref{error}).
           \State $j\rightarrow j+1$
    \EndWhile
  \end{algorithmic}
\end{algorithm}

\section{\textit{BO-ML-ROM} for non-intrusive reduced order modeling} \label{FastSVDMLROM}
The DL models used in \textit{BO-ML-ROM} are a CAE for nonlinear dimensionality reduction, an LSTM for time predictions, and an FFNN for parameter regression \cite{drakoulas2023fastsvd}. In this section, the functionality of these DL models is briefly explained. The training procedure of the DL models based on the gathered data is further described. Finally, this section summarizes the operation of the \textit{BO-ML-ROM} framework during the offline (training) and the online (testing) phase.

\subsection{Dimensionality reduction}
Following the application of the BO framework to calculate the basis matrix $\bm{U}$, the high-fidelity solution corresponding to each parameter $\bm{\theta}_j^{tr}$ at time $t_i$, with $j$=$1,\ldots,m$ and $i$=$1,\ldots, N_t$, is linearly projected using the following formula,
\begin{equation}
    \bm{u}_n(t_i; \bm{\theta}_j^{tr}) = \bm{U}^T \bm{u}_h(t_i; \bm{\theta}_j^{tr})
\end{equation}

In a second step, the solution $\bm{u}_n(t_i; \bm{\theta}_j^{tr})$, is reshaped matrix, applying zero padding, if necessary, and is given as input to the CAE, aiming to compute a low-dimensional latent representation. In particular, the CAE is composed of an encoder and a decoder function. The encoder maps the input tensor to a latent vector. This is achieved by using convolutional layers, max-pooling operations, and dense layers combined with non-linear activation functions. In the inverse direction, the decoder, following a mirrored architecture, reconstructs the reduced solution $\bm{\tilde{u}}_n$ using the latent representation through dense, convolutional, and upsampling layers \cite{goodfellow2016deep}.

Let $\mathbf{\tilde{{z}}}_q(t_i;\bm{\theta}_j^{tr})$ $\in$ $\mathbb{R}^q$, with $q\ll n$ denote the latent vector representation of the input data $\bm{u_n}(t_i;\bm{\theta}_j^{tr})$. Then, the encoder $\bm{\phi}_e: {\mathbb{R}^n} \rightarrow {\mathbb{R}^q}$ is the following mapping,
\begin{equation}
     \mathbf{\Tilde{{z}}}_{\bm{q}}(t_i;\bm{\theta}^{tr}_j)=\bm{\phi}_e(\bm{u}_n(t_i;\bm{\theta}^{tr}_j);\bm{w}_e , \bm{b}_e), \:\:\:  with \:\:\: i=1,\ldots,N_t \:\:\: and \:\:\: j=1,\ldots,m
\end{equation}
where $\bm{w}_e$ and $\bm{b}_e$ are the vectors of weight and bias parameters of the encoder, respectively. The decoder $\bm{\phi}_d: {\mathbb{R}^q} \rightarrow {\mathbb{R}^n}$ approximately reconstructs the linear projected solution through the following transformation,
\begin{equation} \label{Eq:17}
    \bm{\Tilde{u}}_n(t_i;\bm{\theta}^{tr}_j)=\bm{\phi}_d( \mathbf{\Tilde{{z}}}_q(t_i;\bm{\theta}^{tr}_j);\bm{w}_d,\bm{b}_d)=\bm{\phi}_d(\bm{\phi}_e(\bm{u}_n(t_i;{\bm{\theta}^{tr}_j});\bm{w}_e,\bm{b}_e);\bm{w}_d,\bm{b}_d)
\end{equation}
where $\bm{w}_d$ and $\bm{b}_d$ are the vectors of weight and bias parameters of the decoder. For the reconstructed solution, the following equation is utilized,
\begin{equation}
    \bm{\Tilde{u}}_h(t_i;\bm{\theta}^{tr}_j) = \bm{U}\bm{\Tilde{u}}_n(t_i;\bm{\theta}^{tr}_j)
\end{equation}
The CAE is trained by minimizing the mean squared error (mse) loss function $\mathcal{L}^{c}_{mse}$, among the input $\bm{u}_n(t_i;\bm{\theta}^{tr}_j)$, and the reconstructed data $\bm{\Tilde{u}}_n(t_i;\bm{\theta}^{tr}_j)$, with $i=0,\ldots,N_t$ and $j=1,\ldots,m$, given as follows
\begin{equation} \label{Loss_CAE}
    \mathcal{L}^{c}_{mse} = \frac{1}{m \cdot N_t } \sum_{j=1}^{m}\sum_{i=1}^{N_t}\left|\left|\bm{u}_n(t_i;\bm{\theta}_j^{tr})-\bm{\Tilde{u}}_n(t_i;\bm{\theta}_j^{tr})\right|\right|^2
\end{equation}

\subsection{Prediction of dynamics and parameter regression}
In the context of the ROM framework used in this work, the prediction of the dynamics is performed using the low dimensional representation $\mathbf{\Tilde{{z}}}_{{q}}(t_i;\bm{\theta}^{tr}_j)$, obtained through the CAE. In particular, an FFNN is used to build a regression model among the parameter vectors and the latent representations, while an LSTM network, is used to predict the temporal evolution of the latent vectors. The DL models are trained independently, and then during the online phase, the FFNN computes the initial sequence of latent vectors required by the LSTM, to predict the evolution of the latent variables \cite{drakoulas2023fastsvd}. 

 The LSTM network takes as input, vectors of the form $\bm{x}_l^i(\bm{\theta}^{tr}_j)=\icol{\mathbf{\tilde{z}}_q(t_i;\bm{\theta}_j^{tr})\\ \bm{\theta}_j^{tr}}$, consisting of the latent representation, at a specific time instance $t_i$, concatenated with the parameter vector $\bm{\theta}_j^{tr}$. In particular, multiple samples, generated by a predefined sliding window $w$, and arranged in a temporal sequence, enter the LSTM network. For each parameter, the input set has the following form, 
\begin{equation} \label{eq:9}
    \mathcal{X}_l(\bm{\theta}_j^{tr}) = \{ \; \bm{X}_l^1(\bm{\theta}_j^{tr}),\ldots,\bm{X}_l^g(\bm{\theta}_j^{tr}) \; \}
 \end{equation}
where $g=(N_t-w)$, and each matrix $\bm{X}_l^\alpha(\bm{\theta}_j^{tr})$ contains w column vectors, as follows, 
\begin{equation}
    \bm{X}_l^\alpha(\bm{\theta}_j^{tr})=[ \; \bm{x}_l^\alpha(\bm{\theta}_j^{tr}) \;| \ldots| \; \bm{x}_l^{(\alpha+w-1)}(\bm{\theta}_j^{tr})\; ], \: \:  \alpha =1, \ldots, g
\end{equation}
The LSTM network is trained using the batched dataset $\bm{X}_l^\alpha(\bm{\theta}_j^{tr})$ aiming to predict the next latent space vector $\bm{y}_l^\alpha(\bm{\theta}_j^{tr})=\mathbf{\tilde{z}_q}(t_{\alpha+w};  \: \bm{\theta}_j^{tr})$.
The LSTM model is employed using the dataset composed of the set of inputs $\bm{X}_l^\alpha(\bm{\theta}_j^{tr})$, while the output set $\mathcal{Y}_l$ of the training sequence, for each parameter vector $\bm{\theta}_j^{tr}$, has the following form, 
\begin{equation}
    \mathcal{Y}_l(\bm{\theta}_j^{tr})=\{\bm{y}_l^1(\bm{\theta}_j^{tr}),\ldots,\bm{y}_l^g(\bm{\theta}_j^{tr}) \}
\end{equation}

An LSTM network consists of a number of cells connected in series, while an LSTM cell is composed of three gates acting as filters to the input data, namely; the forget gate $\bm{f}_a(\bm{\theta}^{tr}_j)$, the input gate $\bm{i}_a(\bm{\theta}^{tr}_j)$ and, the output gate, $\bm{o}_a(\bm{\theta}^{tr}_j)$. Besides this, the LSTM cell by utilizing the update cell state $\bm{C}_a(\bm{\theta}^{tr}_j)$, the hidden state $\bm{h}_a(\bm{\theta}^{tr}_j)$, predicts the final output $\bm{\tilde{y}}_l^a(\bm{\theta}_j^{tr})$. A detailed explanation of the LSTM architecture can be found in \cite{yu2019review, ghojogh2023recurrent}.

We denote as $\bm{\varphi}_l$ the general function of an LSTM network that aims to predict the evolution of the input data $\bm{X}_l^\alpha(\bm{\theta}_j^{tr})$,
\begin{equation} \label{Eq:23}
    \mathbf{\tilde{y}}_l^\alpha(\bm{\theta}_j^{tr}) = \bm{\varphi}_l(\bm{X}_l^\alpha(\bm{\theta}_j^{tr});\bm{w}_l,\bm{b}_l)
\end{equation}
where $\bm{w}_l$, $\bm{b}_l$ are the weight and bias vectors of the LSTM network, with $\alpha=1, \ldots, g$ and $j= 1, \ldots, m$. The loss function $\mathcal{L}^l_{mse}$ of the LSTM network, is given by,
\begin{equation} \label{Loss_LSTM}
    \mathcal{L}^l_{mse} = \frac{1}{g \cdot m } \sum_{j=1}^{m}\sum_{i=1}^{g}\left|\left|\bm{y}_l^a(\bm{\theta}_j^{tr})-\mathbf{\tilde{y}}_l^a(\bm{\theta}_j^{tr})\right|\right|^2
\end{equation}

In the online phase, the LSTM requires as an input the matrix $\mathbf{X}_l^1(\bm{\theta})$, so as to predict the evolution of the latent representation for the entire time range. Therefore, as an intermediate step, an FFNN is employed to build a regression task to relate the parameter vectors $\bm{\theta}$ to the matrix $\mathbf{X}_l^1(\bm{\theta})$. The data for the training of the FFNN is composed of the training parameter vectors $\bm{\theta}^{tr}$ and the latent representations $\tilde{\bm{z}}_q(\bm{\theta}^{tr})$. In particular, to form the input for the FFNN $\bm{x}_f^\beta(\bm{\theta}_j^{tr})$, with $\beta = 1, \ldots, w$, we concatenate the time instances $t_\beta$, with the parameters $\bm{\theta}_j^{tr}$, and thus $\bm{x}_f^\beta(\bm{\theta}_j^{tr})= \icol{t_\beta \\\bm{\theta}_j^{tr}}$. 
Therefore, the training set of the FFNN is formed by the following data,
\begin{equation}
    \bm{x}_f(\bm{\theta}_j^{tr})=\{\bm{x}_f^1(\bm{\theta}_j^{tr}),\ldots,\bm{x}_f^w(\bm{\theta}_j^{tr})\}
\end{equation}
The output of the network for a given parameter $\bm{\theta}_j^{tr}$ is the following set,
\begin{equation}
    \bm{y}_f(\bm{\theta}_j^{tr})=\{\bm{y}_f^1(\bm{\theta}^{tr}_j),\ldots,\bm{y}_f^w(\bm{\theta}^{tr}_j) \}
\end{equation}
where, $\bm{y}_f^\beta(\bm{\theta}^{tr}_j)=\bm{\Tilde{{z}}}_{\bm{q}}(t_\beta;\bm{\theta}_j^{tr})$.
Consequently, the regression task is based on the following nonlinear mapping, 
\begin{equation} \label{Eq:27}
\mathbf{\tilde{y}}^\beta_f(\bm{\theta}_j^{tr})=\bm{\varphi}_f(\bm{x}_f^\beta(\bm{\theta}_j^{tr});\bm{w}_f,\bm{b}_f)
\end{equation}
where $\bm{w}_f$, $\bm{b}_f$ contain the weights and bias terms of the network, respectively, and $\bm{\varphi}_f:\mathbb{R}^{\xi+1}$ $\rightarrow$ $\mathbb{R}^q$ denotes the FFNN mapping. The loss function, $\mathcal{L}^f_{mse}$ used in the training of the FFNN, is given as follows, 
\begin{equation} \label{Loss_FFNN}
    \mathcal{L}^f_{mse} = \frac{1}{m \cdot w } \sum_{j=1}^{m}\sum_{\beta=1}^{w}\left|\left|\bm{y}_f^\beta(\bm{\theta}^{tr}_j)-\mathbf{\tilde{y}}^\beta_f(\bm{\theta}_j^{tr})\right|\right|^2
\end{equation} 
\begin{figure}[t]
\centering
\includegraphics[width=1\linewidth]{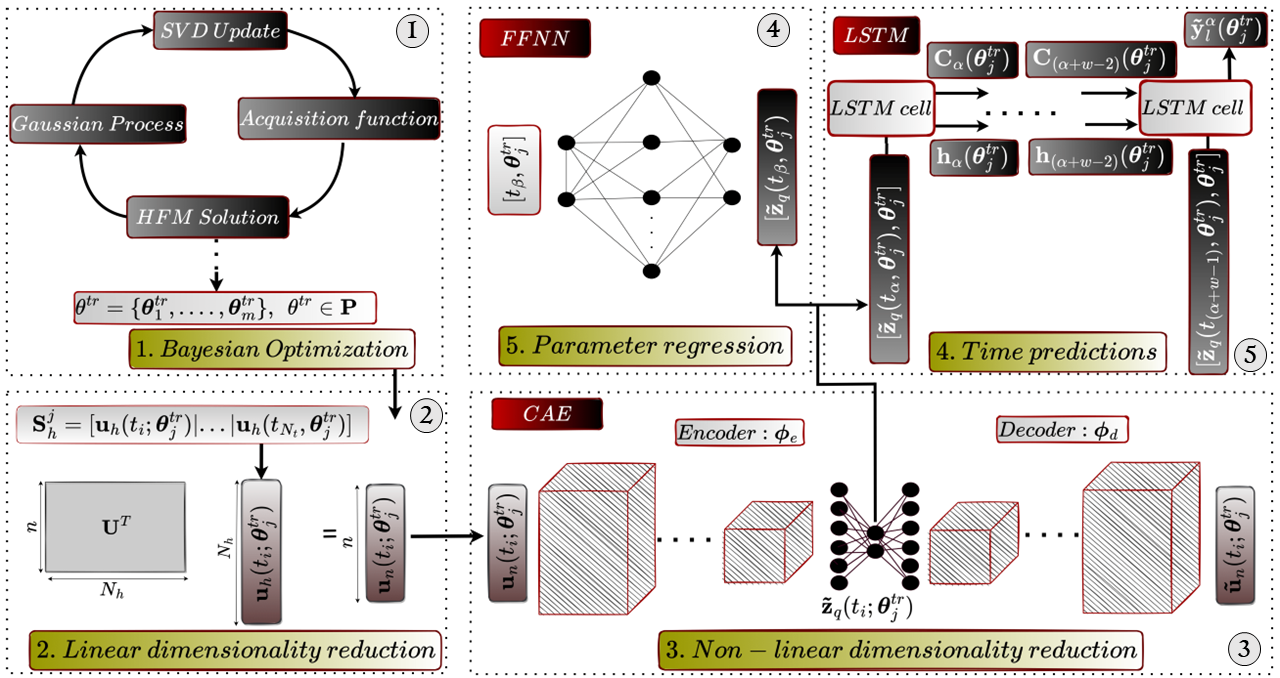}
\caption{Offline phase (training) of the $\textit{BO-ML-ROM}$.} \label{offline fig}
\end{figure}

\subsection{Complete scheme of the \textit{BO-ML-ROM}}
In this subsection, the complete functionality of the \textit{BO-ML-ROM} during the offline and online phases is illustrated. In the offline phase (Fig. \ref{offline fig}), the basis matrix $\bm{U}$ is obtained incrementally, during the simulation process, through an effective combination of the SVD and BO aiming to identify the $m$ most informative training parameter vectors ${\theta}^{tr}$=$\{\bm{\theta}_1^{tr},\ldots,\bm{\theta}_m^{tr}\}$. Once the training dataset of HFM solutions is obtained, the linearly projected data is computed and used for the training of the CAE. Then, the latent representations extracted from the encoder of the CAE, are used for the training of the FFNN for parameter regression, and the training of the LSTM for time predictions.

\begin{figure}[t]
\centering
\includegraphics[width=0.94 \linewidth]{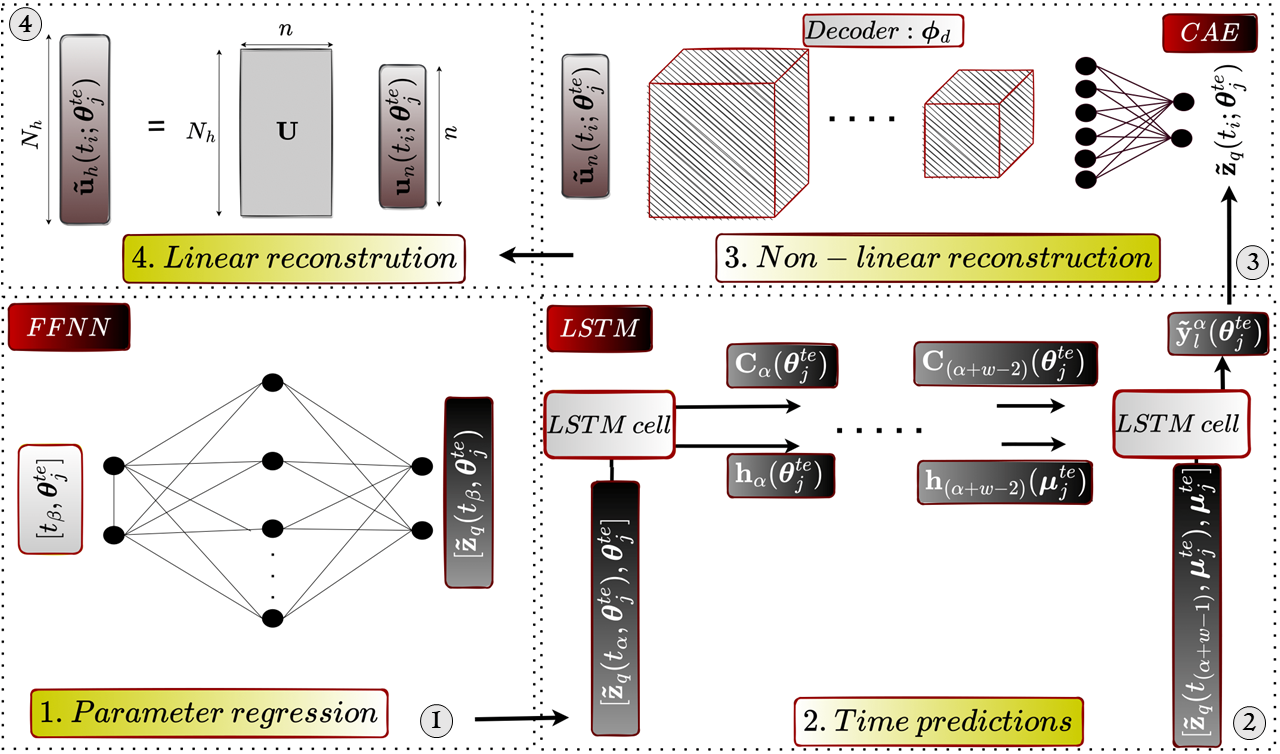}
\caption{Online phase (testing) of the $\textit{BO-ML-ROM}$.} \label{online fig}
\end{figure}

In the online phase, for the testing of the framework, $n$ parameters vectors ${\theta}^{te}$=$\{\bm{\theta}_1^{te},\ldots,\bm{\theta}_n^{te}\}$ are used, randomly sampled from the parameter space $\mathcal{P}$. Each input vector $\bm{\theta}^{te}_j$, with, $j=1,\ldots, n$ is given as input to the trained FFNN together with the first $w$ time instances, to compute the first time sequence of the latent variables and forward the data to the LSTM network that predicts the evolution of the latent variables. Finally, the decoding part of the CAE and the basis matrix $\bm{U}$ are employed to calculate the approximate solutions in a limited amount of time (Fig. \ref{online fig}).  

The accuracy of the ROM predictions for each timestep is assessed using the normalized root mean squared error, $\varepsilon_{rmse} \in \mathbb{R}$ for the testing parameter $\bm{\theta}^{te}_j$, defined as,
\begin{equation}
        {\varepsilon_{nrme}} (t_i) = 
        \frac{ \sqrt{\frac{\sum_{i=1}^{N_h}{ \left (\bm{u}_h(t_i;\bm{\theta}_j^{te})-\mathbf{\tilde{u}}_h(t_i;\bm{\theta}_j^{te})\right)}^2}{N_h}} }{max \left(\bm{u}_h( t_i; \bm{\theta}_j^{te})\right)-min \left(\bm{u}_h(t_i;\bm{\theta}_j^{te})\right)}
    \end{equation}
where $max\left(\bm{u}_h( t_i ;\bm{\theta}_j^{te})\right)$ and $min\left(\bm{u}_h( t_i; \bm{\theta}_j^{te})\right)$, denote the maximum and minimum values of the solution vector $\bm{u}_h(t_i ;\bm{\theta}_j^{te})$.

To improve the performance of the DL models during training, data normalization is applied \cite{drakoulas2023fastsvd}. The minimization of the loss function of each DL model (Eqs. \ref{Loss_CAE}, \ref{Loss_LSTM}, \ref{Loss_FFNN}), is achieved through the backpropagation algorithm, using the adaptive moment estimation optimizer \cite{kingma2014adam}.    

\section{\textit{BO-ML-ROM} for uncertainty quantification (UQ) and sensitivity analysis (SA)} \label{ROMUQ}
In this section, the application of the trained  ROM framework is presented for forward UQ and SA. The former is used to obtain how the uncertainties of the input parameter vectors (e.g., material properties) influence the FEM response, while the latter is utilized to quantitatively estimate the individual or joint contribution of the parameters to the FEM results. For the forward UQ, the Monte Carlo method \cite{hammersley2013monte} is employed, while for the sensitivity analysis, the global, variance-based Sobol' method \cite{saltelli2010variance} is applied.

\subsection{Uncertainty quantification} \label{Ch:UQ}
\begin{figure}[t]
\centering
\includegraphics[width=1\linewidth]{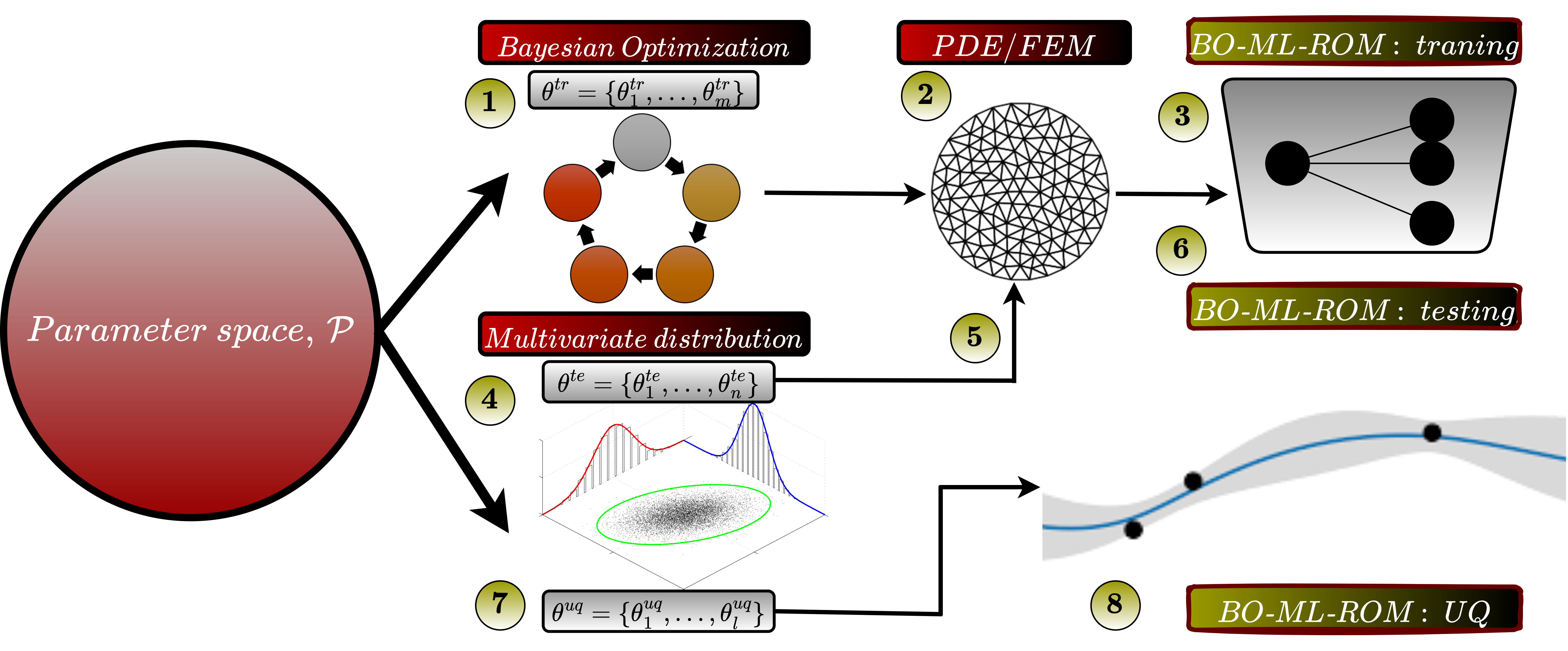}
\caption{The integration of the \textit{FastSVD-ML-ROM} framework with the UQ methodology.} \label{UQ-ROM}
\end{figure}

Once the \textit{BO-ML-ROM} is trained and tested, it is used for forward UQ by replacing the time-consuming FEM solver which application is prohibitive for UQ problems, due to high computation time, related to the numerous required solutions. To compute the effect of the variations in the input parameters (e.g. material properties) on the GWP, we sample from the multivariate Gaussian distribution, $r$ parameter vectors as, $\bm{\theta}^{uq}_j$ $\in\mathcal{P}$ with $j$ = $1,\ldots,r$. For the application of the sampling model, we assume that the input parameters are uncorrelated. The objective is to compute the output field using the \textit{BO-ML-ROM} and study the uncertainty propagation (Fig. \ref{UQ-ROM}). In particular, the \textit{BO-ML-ROM} is employed for the multiple function evaluations required by the Monte Carlo simulation for the forward UQ, overcoming the limitations of this method, related to the increased computational cost, by providing the full-field solution in a limited amount of time \cite{hammersley2013monte}. The Monte Carlo simulation provides the mean $\bm{\tilde{u}}_h^{\mu}(t_j)$ (Eq. \ref{mean}) and standard deviation  $\bm{\tilde{u}}_h^{\sigma}(t_j)$ values (Eq. \ref{std}) at each time instance $t_i$ with $i$ = $1,\dots,N_t$, $\forall$ $\bm{\theta}^{uq}_j$ with $j$ = $1,\ldots,r$, as follows,
\begin{equation} \label{mean}
    \bm{\tilde{u}}_h^{\mu}(t_i) = \frac{1}{r}{\sum_{j=1}^{r}\bm{\tilde{u}}_h(t_i ;\bm{\theta}_j^{uq})}   
\end{equation}
\begin{equation} \label{std}
    \bm{\tilde{u}}_h^{\sigma}(t_i) =  \sqrt{\frac{\sum_{j=1}^{r}\left(\bm{\tilde{u}}_h(t_i ; \bm{\theta}_j^{uq})-\bm{\tilde{u}}_h^{\mu}(t_i)\right)^2}{r-1}}
\end{equation}

\begin{comment}
\textcolor{red}{Then, the calculation of Eq. \ref{uq_menastd} follows, for each time instance $t_i$, to quantify the uncertainties. 
\begin{equation} \label{uq_menastd}
        \bm{\tilde{u}}_h(t_i,\bm{\theta}^{uq})_{uq} =  \bm{\tilde{u}}_h(t_i,\bm{\theta}^{uq})_{mean} \pm \bm{\tilde{u}}_h(t_i,\bm{\theta}^{uq})_{std}
\end{equation}}
\end{comment}
The overall approach, including the implementation of the \textit{BO-ML-ROM} for UQ, followed in this work is depicted in Fig. \ref{UQ-ROM}. The indicative damage indices (DIs) proposed in \cite{ahmed2019uncertainty} are calculated in a second step. DIs are commonly used in the SHM community, since they are sensitive to external factors, i.e. temperature effects, and can estimate the impact of slight variations in the surrounding environment and the structure on the quantities of interest \cite{qiu2016crack}. For the DIs calculation, a normalized root-mean-squared error \cite{barreto2021damage} is defined and computed through the following formula for each parameter value $\bm{\theta}^{uq}_j$,
\begin{equation} \label{DI_eq}
    DI(\bm{\theta}^{uq}_j) = \frac{1}{max\left(DI({\theta}^{uq})\right)}\frac{\sum^{N_t}_{i=1}\Big(\bm{\tilde{u}}_h(t_i; \bm{\theta}^{uq}_j)-\bm{\tilde{u}}_h^{\mu}(t_i)\Big)^2}{\sum^{N_t}_{i=1}\bm{\tilde{u}}_h^{\mu}(t_i)^2}, \:\:\: j =1,\dots,n
\end{equation}
where $max \left(DI({\theta}^{uq}) \right)$ is the maximum value of the damage indexes related to the material properties and the term $\bm{\tilde{u}}_h^{\mu}(t_i)$ represents the baseline (healthy) signal.

\subsection{Sensitivity analysis}
As a next step, we perform a variance-based sensitivity analysis, to quantify the effect of each individual feature (e.g. material properties) of the parameter vector on the propagation of the guided waves, using the variance-based Sobol' method \cite{saltelli2010variance}. The Sobol' method is a global-based SA, in the sense that it considers the variance decomposition across the whole input space, and thus overcomes the limitations of the local methods (e.g., partial derivatives) that depend on the input sample for which the model is evaluated \cite{tosin2020tutorial}. The utilized SA method decomposes the variance of the predicted solution $\hat{V}(\tilde{\bm{u}}_h)$, into a combination of terms that correspond to each individual uncertain input component  $\theta_{\upsilon}$ of the parameter vector (e.g. Young modulus), with $\upsilon=1, \ldots, \xi$, as well as to higher-order terms considering multiple components of the input parameter vector \cite{ye2022uncertainty}, that is, 

\begin{equation}
    \hat{V}(\tilde{\bm{u}}_h) = \sum_\upsilon \hat{V}_\upsilon + \sum_{\upsilon<a} \hat{V}_{\upsilon a} + \ldots  \hat{V}_{1\ldots \xi}
\end{equation}
where $\hat{V}_\upsilon = \hat{V}_{\theta_{\upsilon}}\left(\mathbb{E}_{\bm{\Theta}_{\sim \upsilon}} (\bm{\tilde{u}}_h|\theta_{\upsilon})\right)$ is the variance based first order effect, of $\theta_{\upsilon}$ on the model output and denotes the expected reduction in variance that would be obtained if $\theta_{\upsilon}$ is kept fixed. The inner expectation operation is used to calculate the mean of $\tilde{\bm{u}}_h$ over all possible values of the sampled parameters, stored in the matrix $\bm{\Theta}_{\sim \upsilon}$ to perform the SA, when $\theta_{\upsilon}$ is kept fixed. The outer variance is calculated over all possible values of $\theta_\upsilon$. The partial variance, generated by the interactions among two features $\theta_{\upsilon}$ and $\theta_{a}$, is computed using the relation $\hat{V}_{\upsilon a} =\hat{V}_{\theta_{\upsilon} \theta_{a}}\left(\mathbb{E}_{\bm{\Theta}_{\sim \upsilon a}} (\bm{\tilde{u}}_h|\theta_{\upsilon} \theta_{a})\right) - \hat{V}_{\theta_{\upsilon}}\left(\mathbb{E}_{\bm{\Theta}_{\sim \upsilon}} (\bm{\tilde{u}}_h|\theta_{\upsilon})\right) - \hat{V}_{\theta_{a}}\left(\mathbb{E}_{\bm{\Theta}_{\sim a}} (\bm{\tilde{u}}_h|\theta_{a})\right)$. Similar equations can be derived for the remaining higher order terms \cite{saltelli2010variance}.

\begin{comment}
    The following assumption leads to the decomposition of the variance for each time instance $t_i$ as follows,

\begin{equation} \label{var}
Var \: ( \bm{\tilde{u}}_h(t_j)) = \sum_\kappa V_\kappa \sum_\kappa \sum_{\lambda>\kappa} V_{\kappa \lambda}+ \ldots + V_{12\ldots d}    
\end{equation}
where $V_\kappa , V_{\kappa \lambda}, V_{12\ldots d}$ denote the partial variance contributed by the \textit{$\kappa$}th input, by the interaction between the \textit{$\kappa$}th and the \textit{$\lambda$}th, and by higher order interactions, respectively.
\end{comment}
In the frame of this work, the first-order and total-effect Sobol' indices are computed. The associated sensitivity indicates the first order Sobol' indices $S_\upsilon(t_j)$, of the individual effect of a single feature of the parameter vector $\theta_{\upsilon}$, to the quantity of interest $\tilde{\bm{u}}_h$, for each timestep $t_i$ with $i=1, \ldots, N_t$, as follows \cite{salvador2023fast},
\begin{equation} \label{Si}
    S_\upsilon(t_i) =  \frac{\hat{V}_\upsilon}{\hat{V}(\tilde{\bm{u}}_h)} = \frac{\hat{V}_{\theta_{\upsilon}}\left(\mathbb{E}_{\bm{\Theta}_{\sim \upsilon}} (\bm{\tilde{u}}_h|\theta_{\upsilon})\right)}{\hat{V}(\tilde{\bm{u}}_h)}
\end{equation}
In Eq. \ref{Si}, $S_\upsilon(t_i)$ is a normalized index, since $\hat{V}_{\theta_{\upsilon}}(\mathbb{E}_{\bm{\Theta}_{\sim \upsilon}} (\bm{\tilde{u}}_h|\theta_{\upsilon}))$ varies among zero and $\hat{V}(\tilde{\bm{u}}_h)$, with $\sum_{\upsilon}S_\upsilon(t_i) \leq 1$. The total sensitivity index $S_{T_{\upsilon}}(t_i)$ considers all the contributions of the uncertain feature $\theta_{\upsilon}$, i.e. the first and higher order interactions, at time $t_i$, and is computed as follows,

\begin{equation} \label{ST}
    S_{T_\upsilon}(t_i) = \frac{\mathbb{E}_{\bm{\Theta}_{\sim \upsilon}}\left(\hat{V}_{\theta_{\upsilon}} (\bm{\tilde{u}}_h|\bm{\Theta}_{\sim \upsilon})\right)}{\hat{V}(\tilde{\bm{u}}_h)} = 1 - \frac{\hat{V}_{\bm{\Theta}_{\sim \upsilon}}\Big(\mathbb{E}_{\theta_{\upsilon}} (\bm{\tilde{u}}_h|\bm{\Theta}_{\sim \upsilon})\Big)}{\hat{V}(\tilde{\bm{u}}_h)}
\end{equation}
where, $\sum_{\upsilon}S_{T_\upsilon}(t_i) \geq 1$ since the total effect Sobol' indices, unlike $S_\upsilon(t_i)$, contain also the higher-order terms. For the calculation of the Sobol indices, the \textit{Saltelli} methodology is employed \cite{saltelli2010variance}, using the SALib library \cite{Iwanaga2022}. 

\section{Numerical applications} \label{application}
In the following section, to evaluate the efficiency and the accuracy of the \textit{BO-ML-ROM}, the guided wave propagation in a rectangular aluminum plate is considered. More precisely, UQ and SA are performed with respect to the varying material properties of the structure. For the developed case study, as the quantity of interest, we consider the acceleration field in the excitation direction aligned with the $z$-axis ($\alpha_z$) of a prescribed coordinate system. The FEM is employed for the numerical analysis using the commercial software SIEMENS Simcenter with NX Nastran \cite{anderl2018simulations}, while TensorFlow 2.4 \cite{abadi2016tensorflow} is used for the application of the DL models. The BO algorithm, including the GPR model, is implemented through the Scikit-learn library \cite{sklearn_api}. The numerical simulations and the training of the DL models are executed on an Intel Core i7 @ 2.20 GHz, NVIDIA GeForce GTX 1050 Ti GPU computer.

\subsection{Aluminum plate}
\subsubsection{Numerical model}
In this numerical problem, we consider a 3D linear GWP problem, in an aluminum plate. Our aim is to perform UQ, to evaluate the impact of the variations in the material properties (Young's modulus, Poisson's ratio, density) on the propagation of the waves. These variations are mainly due to the temperature perturbations in the surrounding environment, and the effects of the manufacturing processes of the plate \cite{ahmed2019uncertainty}. Therefore, we define the parameter vector as $\bm{\theta} = [{E}, {\nu}, {\rho}, T]$. In particular, we consider the four components of the parameter vector to be Gaussian random variables, and further these variables are assumed to be uncorrelated. More specifically, four Gaussian distributions are defined for the variations in the material properties and the temperature (Table \ref{tab:mate_all}). 

\begin{figure}[b!]
\centering
\includegraphics[width=0.83\linewidth]{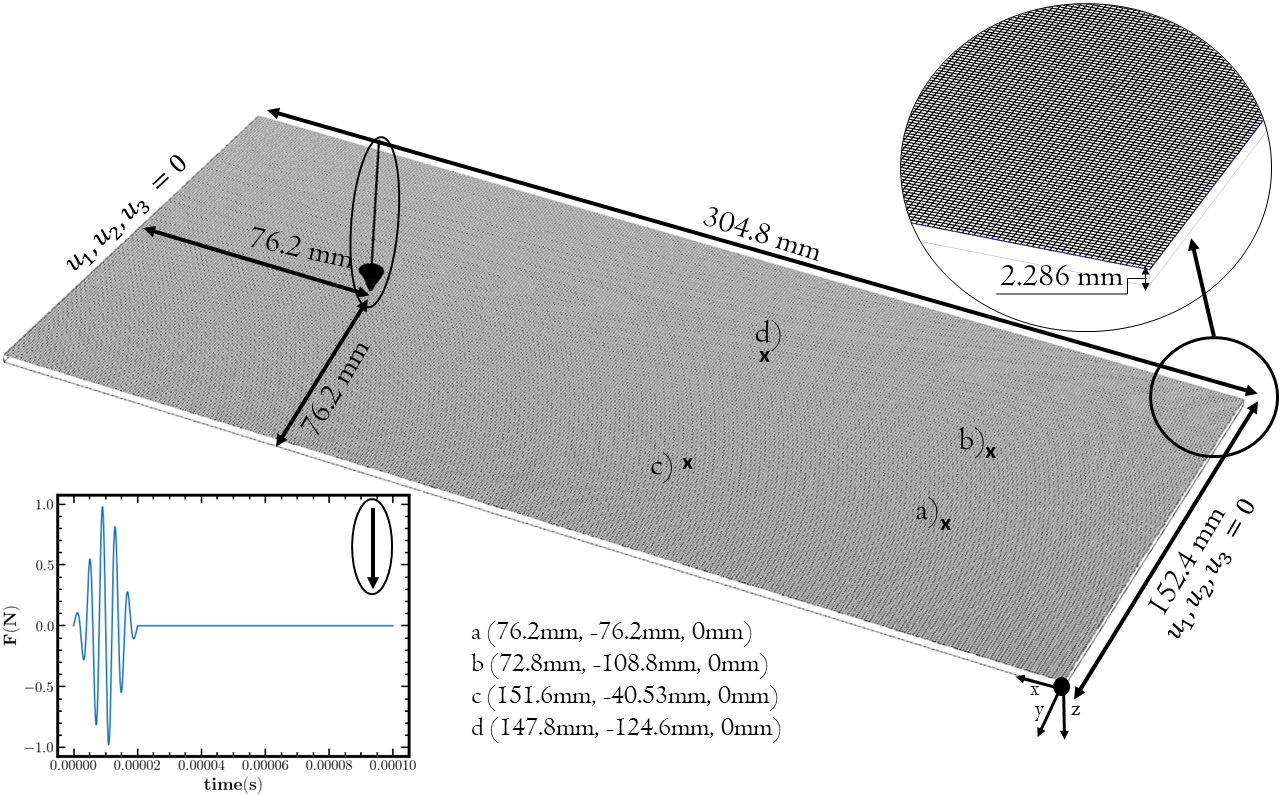}
\caption{Geometry and mesh grid of the rectangular plate, including the boundary conditions.} \label{allum:geometry}
\end{figure}

\begin{table}[h]
\centering{
\caption{Normal distribution for the material properties and temperature.}
\label{tab:mate_all}
\begin{tabular}{|l|c|c|c|c|}
\hline
\textbf{Material} & \begin{tabular}[c]{@{}c@{}}$\mathcal{N}_{E}$\\ $(GPa) $ \end{tabular} & \begin{tabular}[c]{@{}c@{}}$\mathcal{N}_{\nu}$\\ $-$ \end{tabular}  & \begin{tabular}[c]{@{}c@{}}$\mathcal{N}_{\rho}$\\ $(kg/m^3)$ \end{tabular} &  \begin{tabular}[c]{@{}c@{}}$\mathcal{N}_{T}$\\ $(^\circ C)$ \end{tabular}  \\ \hline
      Hexcel IM7/8552     &  $(68.9, 1.332^2)$ & $(0.33, 0.007^2)$  & $(2700, 2.7^2)$  &  $(25, 6^2)$ \\ \hline
\end{tabular}}
\end{table}
Combining the effects of the manufacturing process with the variation of the environmental conditions, the Young modulus, Poisson ratio, and density of the aluminum plate can be expressed as, ${E}^t = E - 0.0263{T}$, ${\nu}^t = \nu + 0.003{T}$, and ${\rho}^t = \rho - 0.184{T}$, respectively \cite{ahmed2019uncertainty}.

For the simulation of the GWP on the plate, a high-quality, fine mesh grid has been constructed with hexahedral, linear, solid elements (Fig. \ref{allum:geometry}). Three elements are used across the thickness of the plate, aiming to capture the dispersion effects in the GWP accurately. In particular, a global mesh size of 0.75 mm is derived through a mesh convergence study, resulting in 328,112 nodes. A five-peak tone burst signal with a central frequency of 250 kHz is used to excite the plate in the $z$-axis direction of the upper face, while two of the plate's faces of the plate are kept fixed as shown in Fig. \ref{allum:geometry}. In terms of the solution setup for the wave propagation, a total time window $T=1 \times 10^{-4}$ s is defined with a timestep $dt=1\times 10^{-7} $ s, yielding 1000 time solutions. We then keep 200 solutions, starting from the first solution with a step equal to 5. To construct the database for the ROM training, the upper face of the plate is used, containing 82,028 nodes. 
\begin{figure}[b!] 
\centering
\includegraphics[width=1 \linewidth]{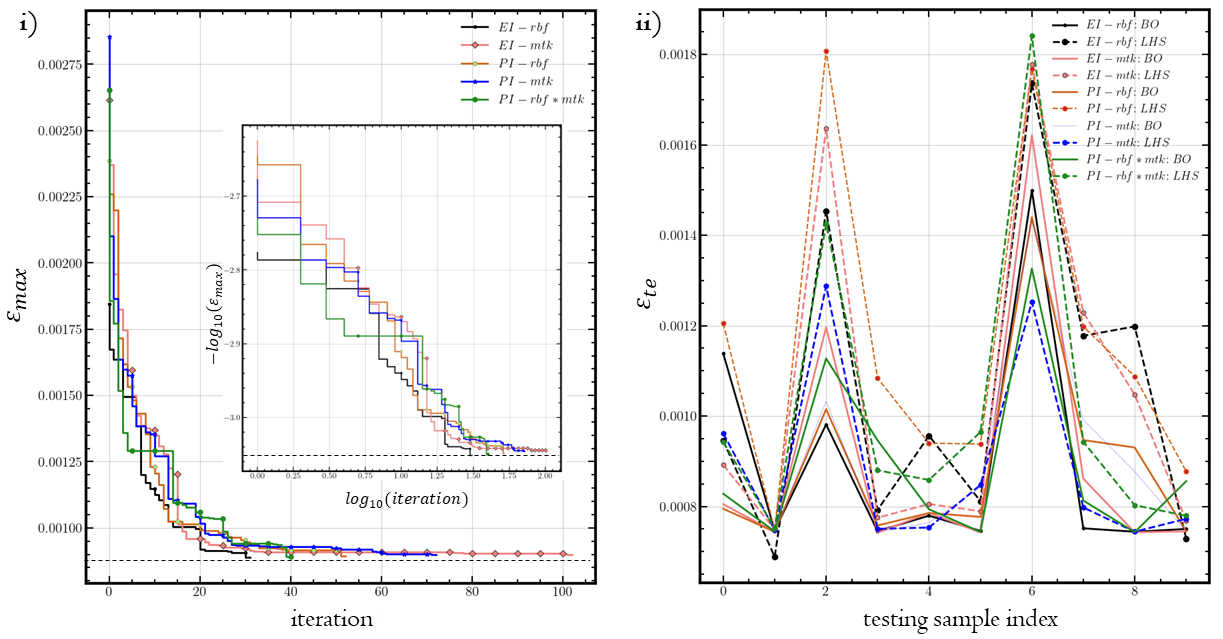}
\caption{i) Maximum reconstruction error evolution across BO iterations for each utilized kernel and AF, ii) reconstruction error for each testing parameter vector index using the different BO setups and the LHS method.} \label{BO_all}
\end{figure}

\subsubsection{\textit{BO-ML-ROM} results}
As a first step for the application of the \textit{BO-ML-ROM}, $\tau=4$ initial parameter vectors are selected from the given multivariate normal distribution, aiming to initialize the BO algorithm. Then, the HFM solutions are calculated using the FEM for these $\tau$ parameter vectors, the basis matrix $\bm{U}_\tau$ is assembled using the SVD update algorithm, and the errors $\varepsilon_j^{tr}$, with $j=1,\ldots, \tau$, related to the reconstructed solutions are computed. The accuracy of the SVD update algorithm is defined to be $\varepsilon_{svd} = 7 \times 10^{-4}$. We then randomly sample $n=10$ testing parameters $\bm{\theta}^{te}$, to determine the convergence and the efficiency of the BO framework compared to LHS, as well as, to test the accuracy of the \textit{BO-ML-ROM}. A stopping error $\varepsilon_{tol} = 9 \times 10^{-4}$ is defined for the BO. In the frame of this work, both $\varepsilon_{svd}$ and $\varepsilon_{tol}$ are selected in an ad-hoc manner. Moreover, multiple combinations of kernels and acquisition functions are examined to optimize the effectiveness of the proposed methodology. While varying the kernels and the AFs, the initial and testing dataset, as well as, the stopping error and truncation error of the SVD update, are kept constant. The objective is to choose the BO setup, that requires the least number of training parameters, and at the same time has the best performance when comparing it with the LHS, for the testing parameters. For the comparison with the LHS, we use the number of training parameter vectors derived from the application of the BO algorithm and compute the basis matrix.  

In Fig. \ref{BO_all}, the results of the BO algorithm for the different utilized AFs and kernels are presented. In particular, Fig. \ref{BO_all}. (i) illustrates the evolution of the $\varepsilon_{max}$ at each iteration. It is deduced that the BO algorithm with the $EI-rbf$ setup requires fewer iterations to reach the stopping error $\varepsilon_{tol}$. Furthermore, the $EI-mtk$ configuration requires the highest number of iterations to converge to $\varepsilon_{tol}$, compared to the other configurations. In Fig. \ref{BO_all}. (ii), we examine the reconstruction error related to the basis matrix $\bm{U}$ for each utilized kernel and AF, and compare the resulting error, of each setup with the obtained error when applying the LHS method, for the testing parameters. It is shown that the $EI-rbf$ using the BO setup leads to the minimum errors $\varepsilon_j^{te}$ with $j=1,\ldots,10$. As it is evident, the BO optimization procedure outperforms the LHS, in terms of the quality of the sampled parameter vectors, leading to higher prediction accuracy, with less training data. 

\begin{comment}
    In particular, the generated errors are: \textit{i)} $EI-RBF$: $\varepsilon_{sum}^{BO}=0.0085$ and $\varepsilon_{sum}^{LHS}=0.0104$, \textit{ii)} $EI-Mtk$: $\varepsilon_{sum}^{BO}=0.0089$ and $\varepsilon_{sum}^{LHS}=0.0106$, \textit{iii)} $PoI-RBF$: $\varepsilon_{sum}^{BO}=0.0088$ and $\varepsilon_{sum}^{LHS}=0.011$, \textit{iv)} $PoI-Mtk$: $\varepsilon_{sum}^{BO}=0.009$ and $\varepsilon_{sum}^{LHS}=0.0092$, \textit{vi)} $PoI-Mtk$: $\varepsilon_{sum}^{BO}=0.0088$ and $\varepsilon_{sum}^{LHS}=0.01$. 
\end{comment}  

\begin{figure}[b!] 
\centering
\includegraphics[width=0.91 \linewidth]{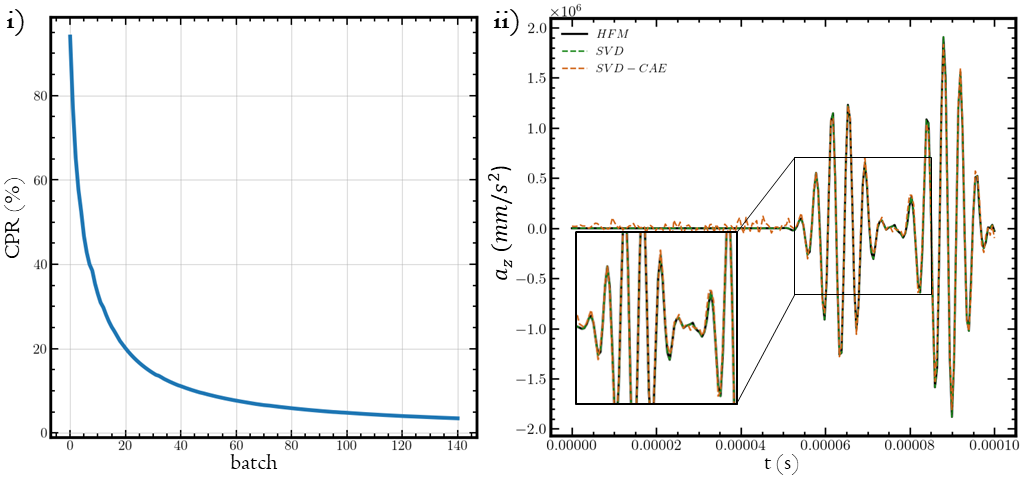}
\caption{i) Evolution of the compression ratio (CPR) and ii) HFM as well as, reconstructed solution after applying SVD and CAE on node a).} \label{SVD_CAE}
\end{figure}

Moving forward, the $EI-rbf$ setup for the acquisition function and the kernel is chosen. When applying the BO algorithm with this setup, $n=36$ training parameters are selected. Having obtained the basis matrix $\bm{U}$, in a second step, we examine the performance of the complete dimensionality reduction process (linear and non-linear). The rank of the computed basis matrix, $\bm{U}$ given $\varepsilon_{svd}$, is $\textit{rank}\:(\bm{U})=249$. In Fig. \ref{SVD_CAE}. (i), the compression ratio (CPR) for the utilized BO setup is presented as a function of the sequential partitions used in the SVD update algorithm \cite{drakoulas2023fastsvd}. The CPR is defined as the fraction of the reduced basis dimension to the full rank of the computed HFMs. In particular, to calculate the CPR, we split each HFM solution into four batches. It is shown that the CPR decays rapidly, exhibiting an approximately exponential behavior, reaching a minimum value equal to $3\%$. Then, to be able to reshape each linear projected solution $\tilde{\bm{u}}_n$ into a square matrix, zero padding is applied to the basis matrix, and each rank is increased into $\textit{rank}\:(\bm{U})=256$. 

In the following, the linear projected data is used for training the DL models. A sequential training methodology is implemented for the DL models to better control the accuracy of the final ROM prediction. Regarding the CAE, we set a latent space dimension $q = 5$, chosen after a hyperparameter analysis. In Fig. \ref{SVD_CAE}, (ii), to investigate the effect of the linear and non-linear dimensionality reduction technique, for a testing HFM solution, in a specific node a) (as shown in Fig. \ref{allum:geometry}). The testing solution corresponds to the parameter vector $\bm{\theta}_4^{te}=[68, 2691, 0.32, 33.1]$. It is shown that both the basis, obtained with the SVD and the CAE, can accurately reconstruct the solution. The CAE reconstruction results in slight noisy perturbations for $t\in[0, 5 \times 10^{-5}]$ s.     

\begin{figure}[h!]
    \centering
\includegraphics[width=0.99 \linewidth]{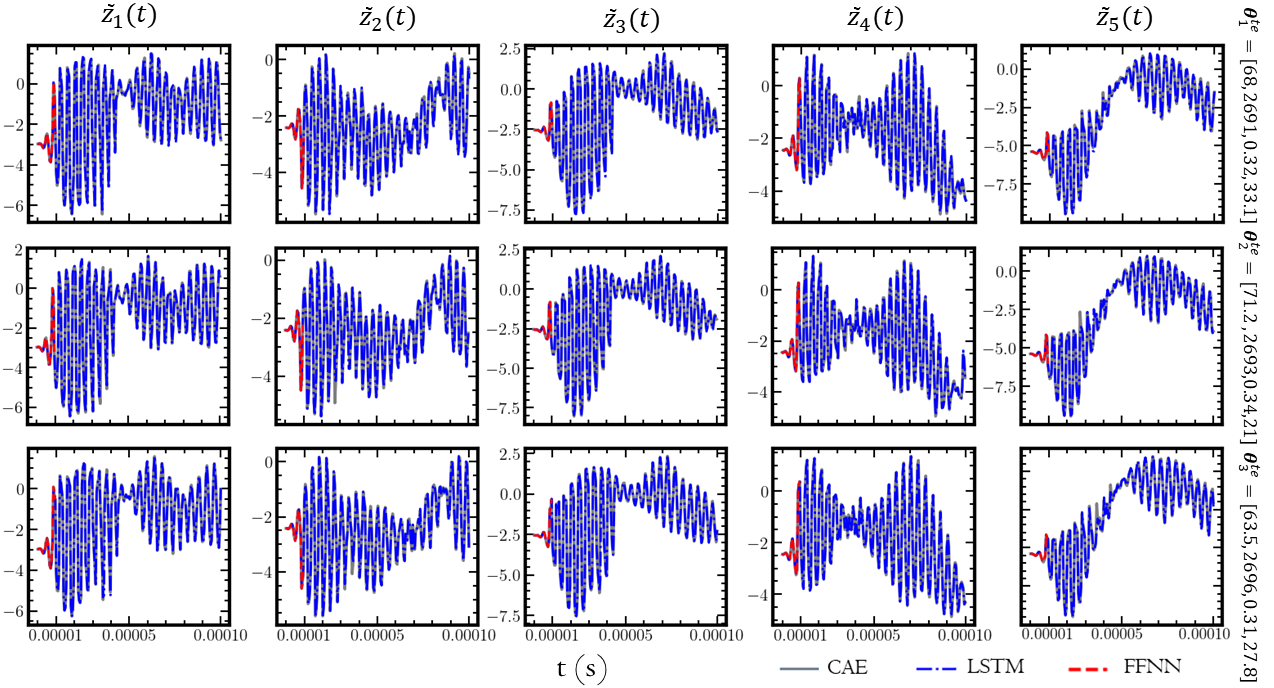}
    \caption{Reference latent variables extracted by the linear basis and the trained encoder CAE, using the HFM solutions at $t=[0, 1 \times 10^{-4}]$ s, FFNN predicted latent variables at $t=[0,2 \times 10^{-6}]$ s and LSTM predicted latent variables at $t=[2 \times 10^{-6}, 1 \times 10^{-4}]$ s for the testing parameters $\bm{\theta}_1^{te}$, $\bm{\theta}_2^{te}$ and $\bm{\theta}_3^{te}$.}
    \label{allum:lstm_ffnn}
\end{figure}

In the next step, the latent spaces extracted from the encoder are used for the LSTM and the FFNN training. Regarding the LSTM, to predict the temporal evolution of the GWP on the aluminum plate using the LSTM network, a time window, $w=10$ is defined. The NN hyperparameters, as well as the NN architectures, are shown in Appendix \ref{NNH}. In particular, Table \ref{NNH: Allum}, presents the NN configurations and parameters, as well as the efficiency of each DL model, including the training and validation losses, and the computation time. To gain a better insight into the ability of the ROM framework to predict the dynamics, we test the performance of the FFNN combined with the LSTM (Fig. \ref{allum:lstm_ffnn}). To achieve that, we compare the results of the extracted latent variables from the trained encoder (reference), with the LSTM predictions for the testing parameters $\bm{\theta}_j^{te}, j=1,2,3$. More specifically, we compute the HFM solutions for $\bm{\theta}_1^{te}$, $\bm{\theta}_2^{te}$ and $\bm{\theta}_3^{te}$, project them into the linear subspace, and forward them to the trained encoder of the CAE, aiming to extract the reference data. Then, we utilize the FFNN to compute the latent variables at $t=[0,2 \times 10^{-6}]$ s, required by the LSTM to predict the latent variables at $t=[2 \times 10^{-6}, 1 \times 10^{-4}]$ s. We observe that the FFNN can accurately predict the first $w$ components of the latent space vectors $\bm{\tilde{z}}$ corresponding to the time interval $t= [0, 2 \times 10^{-6}]$ s. Moreover, we obtain that the LSTM can accurately predict the dynamics in the time range $t=[2 \times 10^{-6}, 1 \times 10^{-4}]$ s. A detailed presentation of the utilized neural network architectures of the FFNN, the CAE, and the LSTM is presented in Appendix \ref{Neural network arctitectures}.

\begin{figure}[t!]
    \centering
\includegraphics[width=0.94 \linewidth]{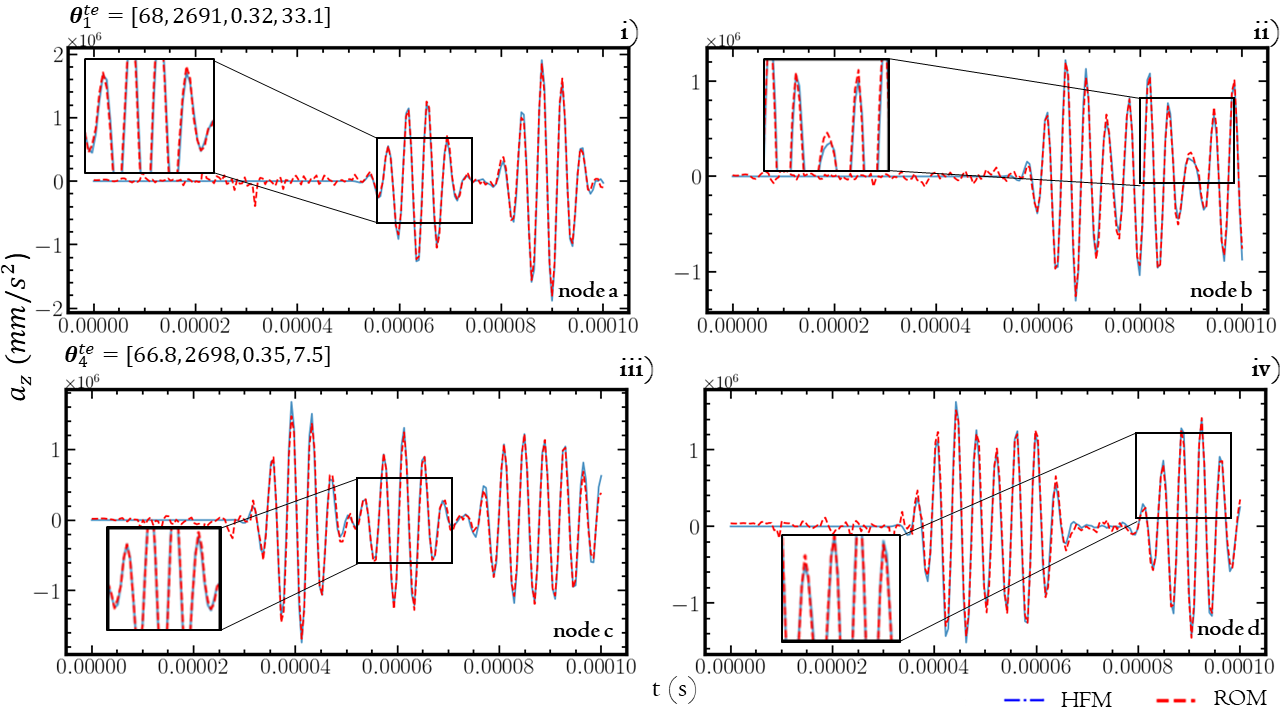}
        \caption{Acceleration as a function of time in nodes a, b, c and d for the parameter vector $\bm{\theta}_1^{te}$ and $\bm{\theta}_4^{te}$.}
    \label{allum:1D_plot}
\end{figure}

Having trained all the DL models, in Fig. \ref{allum:1D_plot}, we test the ability of the proposed ROM framework, to monitor the evolution of the acceleration $a_z$ for the nodes $a$ and $b$ (as shown in Fig. \ref{allum:geometry}) in the time range $t = [0, 1 \times 10^{-4}]$ s. It is obvious that the \textit{BO-ML-ROM} can accurately capture the propagating modes for both testing parameters $\bm{\theta}_1^{te}$ and $\bm{\theta}_2^{te}$, presenting slight noise perturbations, generated through the decoder reconstruction as shown previously.

\begin{figure}[b!]
    \centering
\includegraphics[width=0.94 \linewidth]{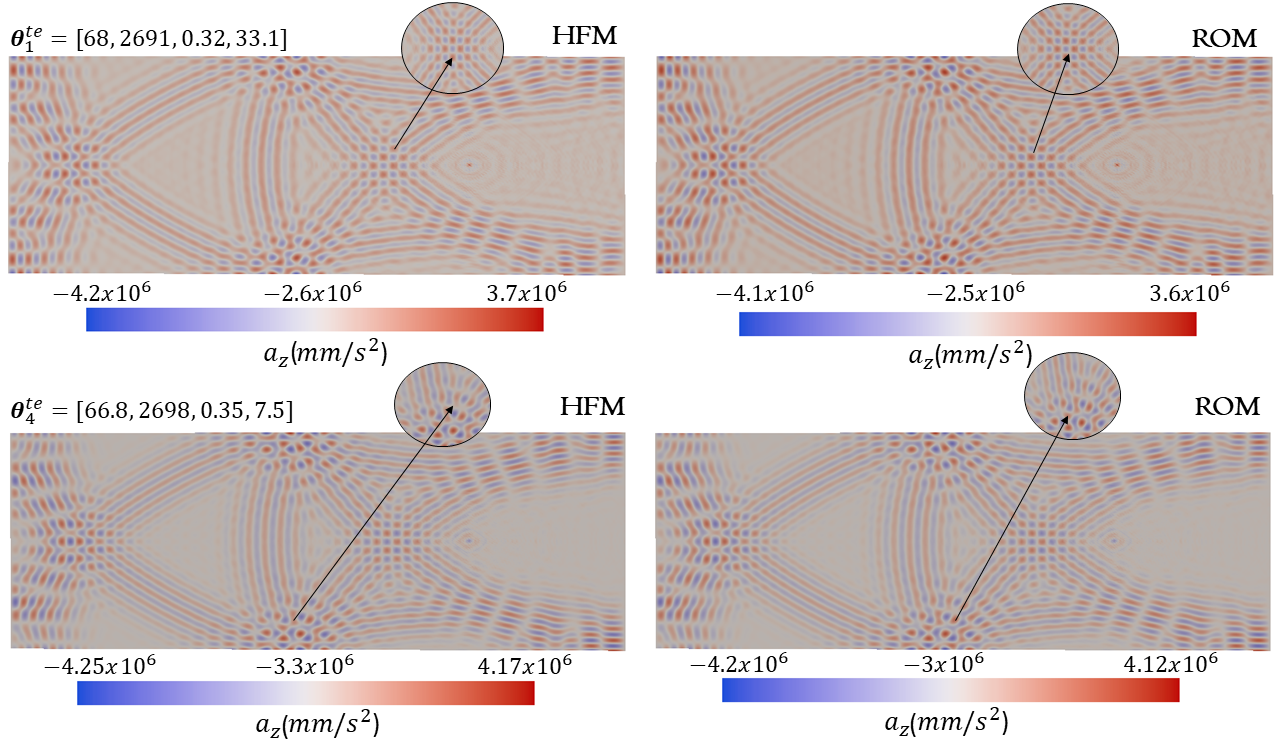}
    \caption{Comparison of the HFM solution to the ROM prediction of the parameter vector $\bm{\theta}_1^{te}$ (upper) and $\bm{\theta}_4^{te}$ (bottom) at time $t= 1 \times 10^{-4}$ s.}
    \label{allum:fom_plate}
\end{figure}
To further assess the performance of the \textit{BO-ML-ROM}, in Fig. \ref{allum:fom_plate}, the results of the true and ROM predicted solutions are presented for $\bm{\theta}^{te}_1$ and $\bm{\theta}^{te}_4$ at $t = 1 \times 10^{-4}$ s. We note the ability of the developed ROM, to compute the full field of the acceleration on the aluminum plate. In particular, for $\bm{\theta}^{te}_1$, an error $\varepsilon_{rmse}=0.015$ is obtained, while, regarding $\bm{\theta}^{te}_4$ an $\varepsilon_{rmse}=0.012$ is obtained.  

\subsubsection{Uncertainty quantification and sensitivity analysis results}
\begin{figure}[t!]
    \centering
\includegraphics[width=1 \linewidth]{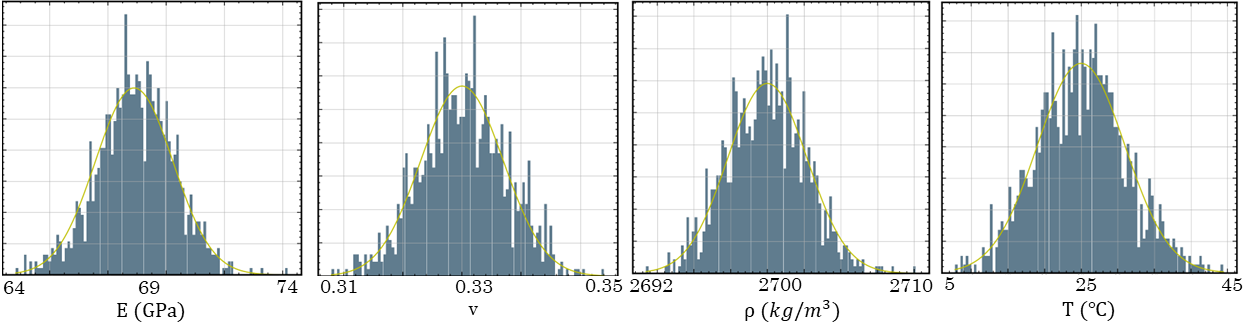}
    \caption{Probability density of each feature used to generate the forward UQ.}
    \label{allum:params_UQ}
\end{figure}

Once the proposed ROM framework is trained and tested, the UQ procedure is implemented (Section \ref{Ch:UQ}) to investigate the effect of the variation in the material properties on the acceleration field. To gain an insight into the forward UQ problem related to the variations of the input features ($E, \nu, \rho, T)$, we sample $r=1000$ parameters vectors (Fig. \ref{allum:params_UQ}). Then, each parameter vector $\bm{\theta}_j^{uq}$ is given as input to the trained \textit{BO-ML-ROM} to compute the acceleration field at each timestep $t_i$, for $i=1,\ldots, 200$. The overall computational time of the UQ procedure is approximately 3 hrs, revealing the dramatic speed-up of the simulation time compared to the utilized FEM solver, which is prohibitive for the solution of 1000 HFMs (requires approximately 1 month).   

\begin{figure}[b!]
    \centering
\includegraphics[width=0.96 \linewidth]{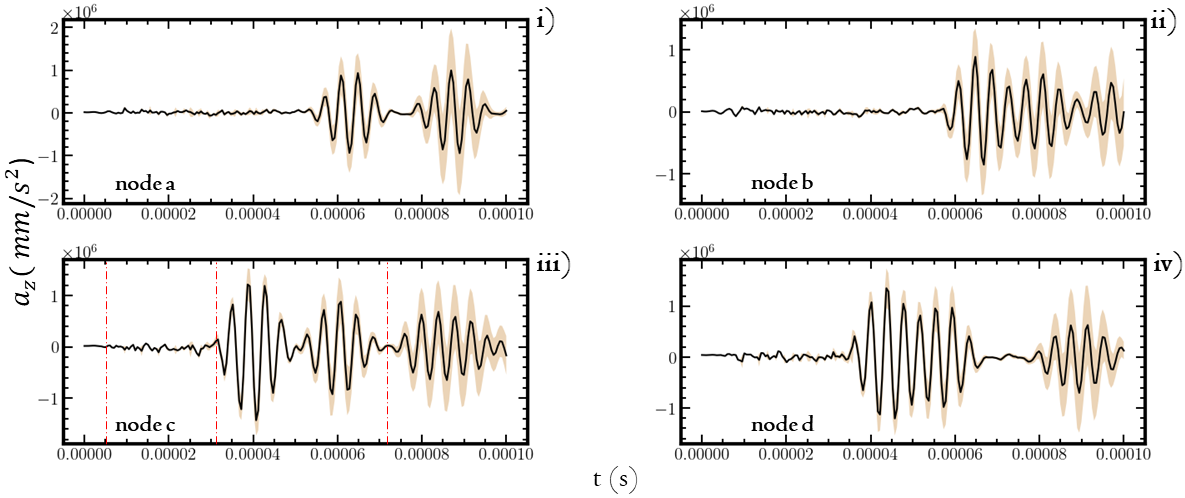}
    \caption{Mean acceleration (black line) and $\pm$ standard deviation (brown area) for the ${\theta}^{uq}$ parameters on nodes $a$, $b$, $c$ and $d$.}
    \label{allum:uq_1d}
\end{figure}

In Fig. \ref{allum:uq_1d}, the acceleration prediction is shown for the parameter vectors $\bm{\theta}^{uq}$ on nodes $a$, $b$, $c$, and $d$ (Fig. \ref{allum:geometry}). More specifically, the black line represents the mean value of the accelerations for all the parameter vectors across time, while the brown area represents the standard deviation. As it is evident, the uncertainties in the material properties significantly affect the acceleration evolution. It is interesting to observe, that the uncertainty increases in the parts of the signals related to the wave reflections. Moreover, it appears that the change in the material properties, mainly affects the amplitude of the signal, while causing much less variation in their phase. 

\begin{figure}[t!]
    \centering
\includegraphics[width=1 \linewidth]{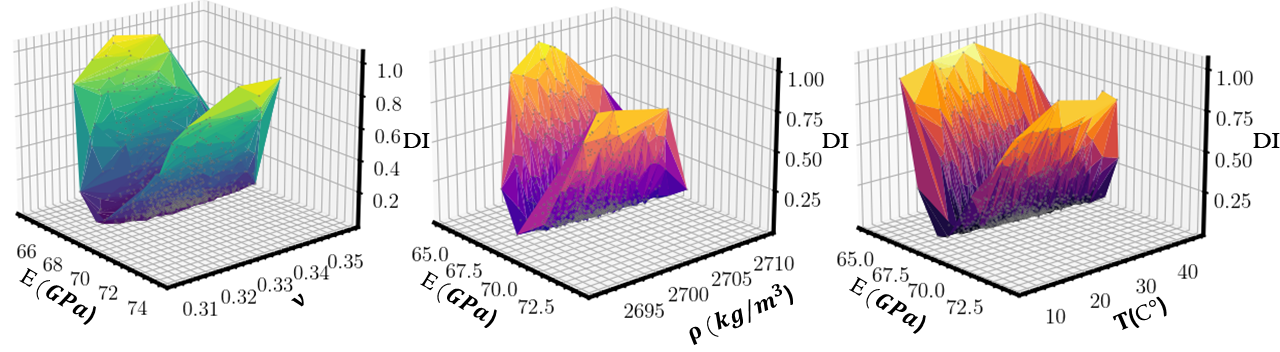}
    \caption{Correlations of the computed damage indices on node $c$, versus the Young modulus ($E$), Poisson ratio ($\rho$), density ($\rho$), and temperature (T).}
    \label{allum:DIs}
\end{figure}

We then calculate the DIs as a function of the material properties on the node $c$. In Fig. \ref{allum:DIs} the computed damage indexes and their variations are presented in terms of the Young modulus, the density, the Poisson ratio, and the temperature. It is shown that the large values of the DIs are obtained in the extremities of each component of the parameter vector. Moreover, the Young modulus ($E$) causes large variations in the DIs values compared to the other components.

\begin{figure}[b!]
    \centering
\includegraphics[width=0.96 \linewidth]{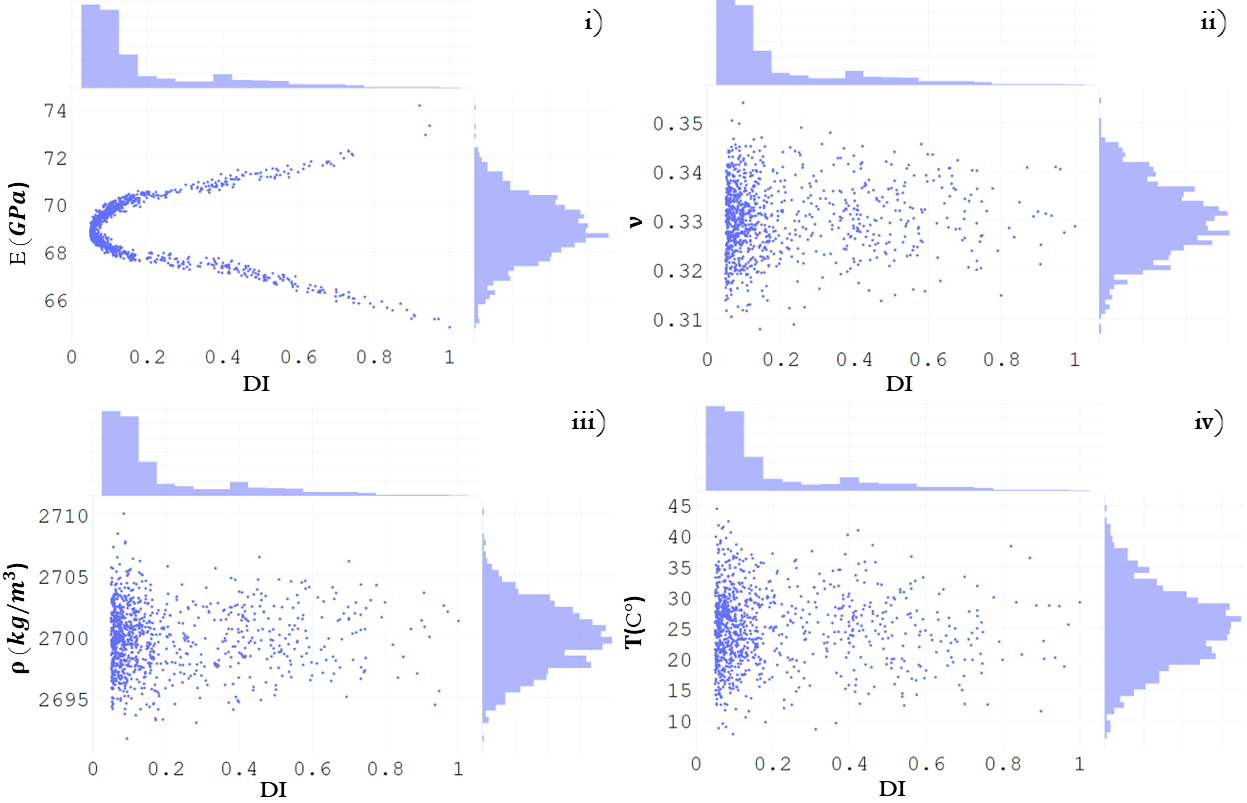}
    \caption{Damage index variation on node $c$, across the young modulus ($E$) and the density ($\rho$).}
    \label{allum:DIs_distr}
\end{figure}

In Fig. \ref{allum:DIs_distr}, we present the correlations of the DI distribution to the material properties distributions. It is observed that the DI distribution presents higher values near the extremities of each component distribution. The lowest values are observed close to the mean value of each distribution.

\begin{figure}[h!]
    \centering
\includegraphics[width=1 \linewidth]{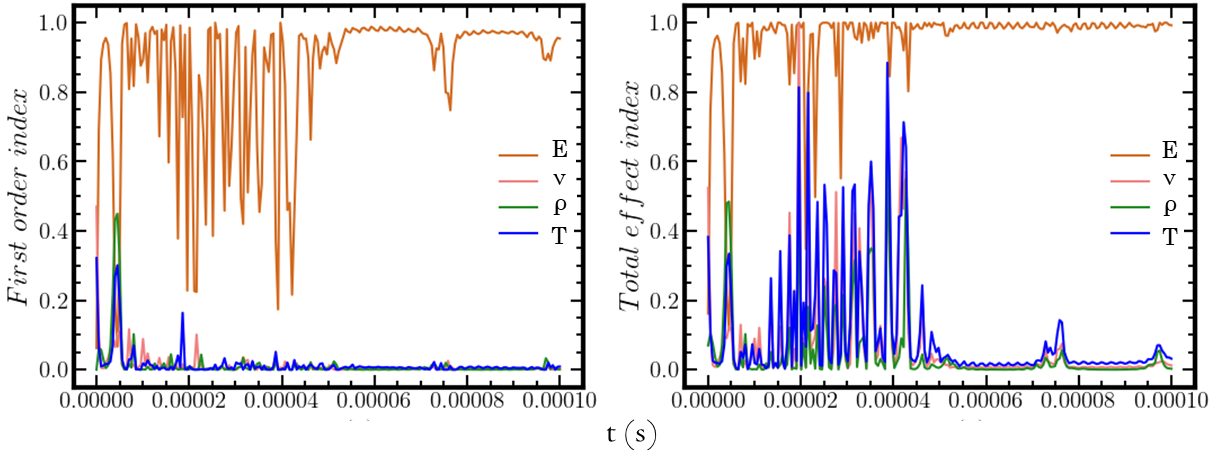}
    \caption{First and total Sobols' indices for the Young modulus $E$, density $\rho$, Poisson ratio $\nu$ and temperature on node $c$, along with time.}
    \label{allum:Sensitivity}
\end{figure}

As a final step, we perform a global, variance-based sensitivity analysis, by computing the first and total Sobol' indices using Saltelli's method (Eqs. \ref{Si}, \ref{ST}) \cite{saltelli2002making}. To accomplish that, following the proposed setup \cite{Iwanaga2022}, we generate $1024(2\xi +2)$ samples, where $\xi$ is the dimension of the input vector, and using the \textit{BO-ML-ROM}, 10,240 solutions are computed, in approximately 10 hours. The objective is to quantify the individual effect of each parametric component on the acceleration field, as well as, the total-effect accounting for the highest interactions among the components of the model. In Fig. \ref{allum:Sensitivity}, it is shown that the Young modulus ($E$) leads to the highest first-order and total effect indices.

\begin{figure}[h!]
    \centering
\includegraphics[width=1 \linewidth]{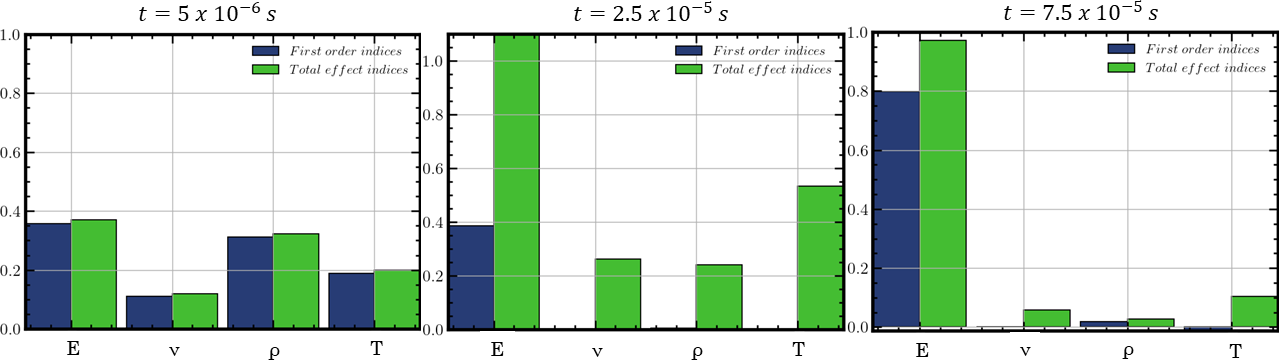}
    \caption{First and total Sobol' indices for the Young modulus $E$, density $\rho$, Poisson ratio $\nu$ and temperature $T$ on node $c$, at t=$5 \times   10^{-6}$ s, t=$2.5 \times 10^{-5}$ s, and t=$7.5 \times 10^{-5}$ s.}
    \label{allum:Sensitivity_params}
\end{figure}

To gain a better insight into the contribution of each parameter to the quantity of interest, we present in Fig. \ref{allum:Sensitivity_params} for node $c$ the first order and total Sobol' indices for each component of the parameter vector at three specific time instances (Fig. \ref{allum:uq_1d}, iii). It is observed that the contribution of each parameter to the Sobol indices varies across time. Moreover, we can deduce that the contribution of the Young modulus remains significant across all the time instances, while the contribution of the other parameters decays with time. 

\section{Conclusions} \label{Conclusions}
In the present work, a non-intrusive ROM framework, referred to as \textit{BO-ML-ROM}, has been developed, for the fast simulation of the guided wave propagation in aluminum plates. The developed ROM is used to perform uncertainty quantification and sensitivity analysis by considering the impact of the varying material properties on the acceleration field. The predicted results imply the effectiveness of the ROM approach, and the comparison between the high-fidelity solutions obtained with the FEM solver, and the ROM predictions, indicates the large speed-up obtained in the calculation of the acceleration field. Moreover, its application for forward UQ and SA reveals the reduced complexity compared to performing the simulations using the FEM solver. It is evident that the ROM can be a powerful tool to evaluate the effects of the variation in the material properties, by computing numerous solutions in a limited amount of time in contrast with the prohibitive computational cost of HFM simulation. The UQ results point out that the material properties significantly affect the acceleration field. In particular, the SA results reveal that the effect in the predicted results caused by the Young modulus is higher than the corresponding effect produced by the Poisson ratio, the density, and the temperature.  

The developed ROM framework utilizes a BO algorithm, to adaptively select the parameter vectors for the ROM training. It is shown that the sequential GPR-based adaptive parameter sampling methodology outperforms the one-shot LHS method in terms of the required number of parameter vectors to reach a predefined accuracy. In particular, the developed algorithm utilizes a stopping criterion, that controls the convergence of the BO algorithm. By using the rbf kernel and the EI acquisition function, the BO method leads to lower errors with respect to the prediction for the testing parameters. 

\begin{comment}
    The \textit{BO-ML-ROM} constitutes a useful tool in the SHM field for non-destructive tests, targeting to improve the monitoring systems for virtual sensing, by enhancing the damage quantification systems.
\end{comment} 
The flexibility and efficiency of the developed platform, are critical aspects for the integration of the ROMs and data-driven methods within the context of SHM, aiming to formulate robust, hybrid, DT systems. Future steps, among others, include the utilization of the sensor measurements for model updating focussing on for the HFM calibration, and the utilization of the ROM for damage identification and localization. 

\section*{Appendix}
\addcontentsline{toc}{section}{Appendices}
\renewcommand{\thesubsection}{\Alph{subsection}}
\subsection{NN-configuration} \label{NNH}
The neural network hyperparameters, utilized in the current work, are presented in this chapter.

\begin{table}[h!]
\caption{NN-configuration and training efficiency for the GWP on the aluminum plate. The computational time of the ROM is compared to the HFM simulation time which requires 45 mins for a given parameter.}
\centering{
\begin{tabular}{|c|c|c|c|}
\hline
\diagbox[innerwidth =40mm, height =15mm]%,innerleftsep = 12pt
{\raisebox{2.5ex}{$\textbf{NN-conf.}$}}{\raisebox{-0.5ex}{$\textbf{DL-model}$}} & $\textbf{CAE}$ & $\textbf{FFNN}$ & $\textbf{LSTM}$ \\ \hline
$$\textbf{learning rate}$$ & $0.0005$        & $0.01$           & $0.0001$         \\ \hline
$\textbf{batch}$         & $20 $           & $4$          & $10$             \\ \hline
$\textbf{epochs}$        & $2,000$         & $10,000$          & $35,000$          \\ \hline
$\textbf{val. loss}$  & $7.4\times10^{-5}$ & $3.9\times10^{-6}$ & $1.1\times10^{-5}$ \\ \hline
$\textbf{train. loss}$ & $9.2\times10^{-4}$ & $5.4\times10^{-6}$ & $3\times10^{-4}$ \\ \hline
$\textbf{training time}$ & $10$ hrs         & $30$ min          & $3$ hrs          \\ \hline
$\textbf{testing time}$  & $0.05$ s       & $0.013$ s       & $0.2$ s         \\ \hline
\end{tabular}} \label{NNH: Allum}
\end{table}

\subsection{Neural network arctitectures} \label{Neural network arctitectures}
The architecture of the neural networks, utilized in the current work, is presented in this chapter.
% FFNN
\begin{table}[H]
\centering{
\begin{tabular}{|c|c|}
\hline
\textbf{Activation function} & Leaky ReLU            \\ \hline
\textbf{Layer}               & \textbf{Output shape} \\ \hline
Input                        & 3                     \\ \hline
Dense                        & 50                    \\ \hline
Dense                        & 4                     \\ \hline
\end{tabular}\caption{FFNN architecture for the GWP on the aluminum plate.}} \label{FFNN: Allum}
\end{table}

%CAE
\begin{table}[H]
\centering{
\begin{tabular}{|cc|cc|}
\hline
\multicolumn{2}{|c|}{\textbf{Encoder}}             & \multicolumn{2}{c|}{\textbf{Decoder}}           \\ \hline
\multicolumn{2}{|c|}{\textbf{Activation function}} & \multicolumn{2}{c|}{ELU}                        \\ \hline
\multicolumn{2}{|c|}{\textbf{Kernel size}}         & \multicolumn{2}{c|}{(3 x 3)}                    \\ \hline
\multicolumn{1}{|c|}{\textbf{Layer}} & \textbf{Output shape} & \multicolumn{1}{c|}{\textbf{Layer}} & \textbf{Output shape} \\ \hline
\multicolumn{1}{|c|}{Input}        & (16,16,1)     & \multicolumn{1}{c|}{Input}       & 4            \\ \hline
\multicolumn{1}{|c|}{Conv2D}       & (16, 16, 25)  & \multicolumn{1}{c|}{Dense}       & 10           \\ \hline
\multicolumn{1}{|c|}{Max pooling}  & (8, 8, 25)    & \multicolumn{1}{c|}{Dense}       & 20           \\ \hline
\multicolumn{1}{|c|}{Conv2D}       & (8, 8, 10)    & \multicolumn{1}{c|}{Dense}       & 12           \\ \hline
\multicolumn{1}{|c|}{Max pooling}    & (4, 4, 10)            & \multicolumn{1}{c|}{Reshape}        & (2, 2, 3)             \\ \hline
\multicolumn{1}{|c|}{Flatten}      & 160           & \multicolumn{1}{c|}{Conv2D}      & (2, 2, 10)   \\ \hline
\multicolumn{1}{|c|}{Dense}        & 20            & \multicolumn{1}{c|}{Up sampling} & (4, 4, 10)   \\ \hline
\multicolumn{1}{|c|}{Dense}        & 10            & \multicolumn{1}{c|}{Conv2D}      & (4, 4, 25)   \\ \hline
\multicolumn{1}{|c|}{Dense}        & 4             & \multicolumn{1}{c|}{Up sampling} & (8, 8, 25)   \\ \hline
\multicolumn{1}{|c|}{-}            & -             & \multicolumn{1}{c|}{Conv2D}      & (8, 8, 30)   \\ \hline
\multicolumn{1}{|c|}{-}            & -             & \multicolumn{1}{c|}{Up sampling} & (16, 16, 30) \\ \hline
\multicolumn{1}{|c|}{-}            & -             & \multicolumn{1}{c|}{Conv2D}      & (16, 16, 1)  \\ \hline
\end{tabular}\caption{CAE architecture for the GWP on the aluminum plate.}}\label{CAE: Allum}
\end{table}
\setlength{\parskip}{0pt}
%LSTM
\begin{table}[H]
\centering{
\begin{tabular}{|c|c|}
\hline
\textbf{Layer} & \textbf{Output shape} \\ \hline
Input          & (10, 6)               \\ \hline
LSTM           & (10, 50)              \\ \hline
LSTM           & (10, 50)              \\ \hline
LSTM           & 50                   \\ \hline
Dense          & 4                     \\ \hline
\end{tabular}\caption{LSTM architecture for the GWP on the aluminum plate.}}\label{LSTM: Allum}
\end{table}

\bibliographystyle{elsarticle-num} 
% elsarticle-harv
\bibliography{references}

%\begin{thebibliography}{00}


\end{document}